\documentclass[twoside]{article}
\usepackage{xcolor}
\RequirePackage{natbib}

\usepackage{tikz-cd}
\usepackage{amsfonts}

\usepackage[linesnumbered,ruled]{algorithm2e}
\SetKwInput{KwInput}{Input}
\SetKwInput{KwOutput}{Output}
\newcommand{\lIfElse}[3]{\lIf{#1}{#2 \textbf{else}~#3}}

\SetCommentSty{mycommfont}
\usepackage{multirow}
\usepackage{booktabs}
\usepackage{hyperref}       
\hypersetup{
    colorlinks=true,
    linkcolor=blue,
    filecolor=magenta,
    urlcolor=cyan,
}

\newenvironment{talign}
 {\align}
 {\endalign}

\usepackage[accepted]{aistats2023}
\def\Pcal{{\mathcal{P}}}

\def\div{{\textrm{div}}}

\DeclareMathOperator*{\argmin}{arg\,min}

\usepackage{amsthm}
\theoremstyle{plain}
\newtheorem{theorem}{Theorem}[section]

\newtheorem{lemma}[theorem]{Lemma}

\theoremstyle{definition}
\newtheorem{definition}[theorem]{Definition}

\newcommand\figref{Figure~\ref}

\begin{document}

%

%
\runningauthor{Haoran Sun, Hanjun Dai, Bo Dai, Haomin Zhou, Dale Schuurmans}

\twocolumn[

\aistatstitle{Discrete Langevin Samplers via Wasserstein Gradient Flow}

\aistatsauthor{Haoran Sun\footnotemark \\
hsun349@gatech.edu 
\And
Hanjun Dai \\
hadai@google.com
\And
Bo Dai \\
bodai@google.com 
}
\aistatsaddress{ Georgia Tech \And Google Research \And Google Research \& Georgia Tech
}
\aistatsauthor{
Haomin Zhou \\
hmzhou@math.gatech.edu
\And
Dale Schuurmans \\
schuurmans@google.com
}
\aistatsaddress{ Georgia Tech \And Google Research \& University of Alberta
}
]

\begin{abstract}
It is known that gradient based MCMC samplers for continuous spaces, such as Langevin Monte Carlo (LMC), can be derived as particle versions of a gradient flow that minimizes KL divergence on a Wasserstein manifold.  The superior efficiency of such samplers has motivated several recent attempts to generalize LMC to discrete spaces.  However, a fully principled extension of Langevin dynamics to discrete spaces has yet to be achieved, due to the lack of well-defined gradients in the sample space.  In this work, we show how the Wasserstein gradient flow can be generalized naturally to discrete spaces.  Given the proposed formulation, we demonstrate how a discrete analogue of Langevin dynamics can subsequently be developed.  With this new understanding, we reveal how recent gradient based samplers in discrete spaces can be obtained as special cases by choosing particular discretizations. More importantly, the framework also allows for the derivation of novel algorithms, one of which, \textit{Discrete Langevin Monte Carlo} (DLMC), is obtained by 
a factorized estimate of the transition matrix.
The DLMC method admits a convenient parallel implementation and time-uniform sampling that achieves larger jump distances.  We demonstrate the advantages of DLMC on various binary and categorical distributions.

\end{abstract}

\section{INTRODUCTION}
The Markov Chain Monte Carlo (MCMC) algorithm is one of the most widely used methods for sampling from intractable distributions~\citep{robert2013monte}. However, it 
is known to mix slowly in complex, high-dimensional models.
In response, 
several gradient based MCMC methods
have been developed over the past decades that leverage gradient information to guide proposals toward high probability regions \citep{neal2011mcmc}.
By simulating Langevin dynamics (LD), the Langevin Monte Carlo method (LMC) \citep{rossky1978brownian} and its variants \citep{welling2011bayesian, girolami2011riemann} have substantially improved  sampling efficiency in both theory and practice.
In seminal work, \citet{jordan1998variational} and \citet{otto2001geometry} have shown that Langevin dynamics simulate $\frac{d}{dt}\rho^t = - \nabla_\rho D_\text{KL}(\rho^t||\pi)$, which is a Wasserstein gradient flow (WGF) that minimizes the KL-divergence to a target distribution $\pi$. This connection not only provides a tool for algorithm design \citep{ma2015complete, liu2019understanding} but also help theoretical analysis \citep{cheng2018convergence}.

\footnotetext{Work done during an internship at Google.}

Despite these advances, progress in gradient based methods has generally focused on continuous spaces. Recently, a family of locally balanced (LB) samplers \citep{zanella2020informed, grathwohl2021oops, sun2021path, zhang2022langevin, sun2022optimal, rhodes2022enhanced} have leveraged gradient information for proposals in discrete spaces via LB functions, 
achieving significant success.
However even though \cite{zanella2020informed} and \citet{sun2021path} have proved that LB functions are asymptotically optimal for leveraging gradient information in a proposal distribution, a principled extension of LMC from continuous to discrete spaces remains lacking
in finite dimensional problems. 
Consequently, existing LB samplers suffer from inefficiencies arising from a suboptimal imitation of LMC. For example, \citet{sun2021path, sun2022optimal} flip multiple sites in order, which prevents parallel implementation; \citet{zhang2022langevin} and \citet{rhodes2022enhanced} restrict a Gaussian proposal to discrete states, which ignores the difference between continuous diffusion and discrete jump processes.

To migrate LMC from continuous to discrete spaces, we consider an alternative perspective that lifts the commonly used particle level to a more principled distribution level view. In particular, we start from the fact that every stochastic process $X^t$ has a corresponding probability density $\rho^t(x)$ in the Wasserstein manifold \citep{villani2009optimal}. Instead of designing jump processes as discrete analogues of the diffusion process in LMC, we instead consider the \textit{Discrete Wasserstein Gradient Flow} (DWGF) $\rho^t$ that minimizes the KL-divergence $D_\text{KL}(\rho^t||\pi)$ to the target distribution $\pi$. We then derive a \textit{Discrete Langevin Dynamics} (DLD) $X^t$ as a particle realization of this gradient flow $\rho^t$.
Unsurprisingly, previous LB samplers can be interpreted as different discretizations of $X^t$, which explains their success in finite dimensional problems. However, the key benefit of this alternative perspective is the development of new algorithms that more faithfully follow the derivation of LMC from LD in continuous spaces. Using a more efficient discretization of DLD, we develop a novel sampler, 
\textit{Discrete Langevin Monte Carlo} (DLMC), that factorizes $X^t$ into sub-processes $X_n^t$, where for each sub-process, the transition $P_n^h$ after a simulation time $h$ is estimated and a new state proposed according to $P_n^h(X^t_n, \cdot)$. In this way, DLMC is (i) computationally more efficient than previous methods, as it decouples dimensions and permits convenient parallel implementation, and (ii) statistically more efficient, since it uses a time-uniform discretization of DLD. The main ideas behind the overall framework can be summarized in the following diagram that relates the derivation of LMC from LD and WGF to their discrete counterparts.
\[
  \begin{tikzcd}[column sep = 8em]
    \arrow[d, "f_2 \text{: Fokker-Planck}"'] \text{WGF} \arrow[r, "f_3 \text{: Section \ref{sec:dfpe}}"]  & \text{DWGF} \arrow[d, "f_4 \text{: Section \ref{sec:dld}}"] \\
     \arrow[d, "f_1 \text{: Discretization}"'] \text{LD}   & \text{DLD} \arrow[d, "f_5 \text{: Section \ref{sec:dlmc}}"] \\
    LMC \arrow[r, "f_5 \circ f_4 \circ f_3 \circ f_2^{-1} \circ f_1^{-1}"] & DLMC
  \end{tikzcd}
\]
An experimental evaluation demonstrates that DLMC enjoys better proposal quality and greater efficiency than traditional samplers as well as other LB samplers.  These advantages are demonstrated both in sampling and learning tasks involving binary and categorical distributions, including the Bernoulli distribution, Ising model, Potts model, factorial hidden Markov model, restricted Boltzmann machine, and deep energy based models (EBMs). Code for reproducing all the experiments can be found at \href{https://github.com/google-research/google-research/tree/master/dwgf}{https://github.com/google-research/google-research/tree/master/dwgf}.

\section{PRELIMINARIES}
We first revisit the framework of WGF $\rightarrow$ LD $\rightarrow$ LMC in the above diagram, which derives LMC as a discretization of LD that minimizes the KL-divergence to a target distribution $\pi$.

\paragraph{Wasserstein Gradient Flow} 
The Wasserstein manifold
\begin{equation}
    \mathcal{P}_2(\mathbb{R}^d) = \{\mu: \mathbb{R}^d \rightarrow \mathbb{R}_{\ge 0}: \int_{\mathbb{R}^d} \mu(x) dx = 1\}
\end{equation}
is the set of probability measures on $\mathbb{R}^d$ where we define the distance between two measures $\mu, \nu \in \mathcal{P}_2(\mathbb{R}^d)$ by the Wasserstein-2 distance \citep{villani2009optimal}: 
\begin{align}
    W_2(\mu, \nu) = \left(\inf_\Pi \int d^2(x, y) d\Pi(x, y)\right)^\frac{1}{2}, \\
    \text{s.t. } \int \Pi(x, y) dy = \mu(x), \int \Pi(x, y) dx = \nu(y), \label{eq:marginal}
\end{align}
where $d(x, y)$ is the distance on the underlying Euclidean space $\mathbb{R}^d$, and $\Pi$ is a joint distribution satisfying the marginal constraints \eqref{eq:marginal}.
For a target distribution $\pi$ and a current distribution $\rho$ on $\mathcal{P}_2(\mathbb{R}^d)$, the gradient of the KL-divergence with respect to $\rho$ is:
\begin{equation}
    \nabla_\rho D_\text{KL}(\rho || \pi) = \nabla \cdot [\log \pi(x) \rho(x)] - \Delta \rho(x).
    \label{eq:wgf}
\end{equation}
A flow $\rho_t$ on $\mathcal{P}_2(V)$ that satisfies
\begin{equation}
    \frac{\partial}{\partial t}\rho^t(x) = - \nabla \cdot [\log \pi(x) \rho^t(x)] + \Delta \rho^t(x), \label{eq:fke}
\end{equation}
is the \textit{Wasserstein gradient flow} (WGF) that minimizes the KL-divergence to the target distribution $\pi$.

\paragraph{Langevin Dynamics} \citet{jordan1998variational} and \citet{otto2001geometry} have established the elegant connection between the WGF \eqref{eq:wgf} and its particle level realization via the Fokker-Planck Equation. In particular, assume $X^t$ is a time dependent random variable that has movement described by the following stochastic differential equation:
\begin{equation}
    d X^t = \nabla \log \pi(X^t) dt + \sqrt{2}dW^t, \label{eq:ld}
\end{equation}
where $W^t \in \mathbb{R}^d$ is a Wiener process.
Then the Fokker-Planck equation asserts that $\rho^t$, the distribution of $X^t$, evolves over time in a way that satisfies \eqref{eq:fke}. The stochastic process in \eqref{eq:ld} is called \textit{Langevin Dynamics} (LD). 


\paragraph{Langevin Monte Carlo} Since LD is a particle realization of the gradient flow \eqref{eq:fke}, one can use it to efficiently generate samples from the target distribution $\pi$ via discrete time simulation:
\begin{equation}
    X^{t+\epsilon} = X^t + \epsilon \nabla \log \pi(X^t) + \sqrt{2\epsilon} \xi, \label{eq:ld_simu}
\end{equation}
where $\xi \sim \mathcal{N}(0, I_d)$ is a standard normally distributed random variable. In practice, the discrete time simulation in \eqref{eq:ld_simu} has an approximation error, so a Metropolis-Hastings acceptance test (MH) is commonly used to correct any bias \citep{metropolis1953equation, hastings1970monte}. Specifically, given the current state $x$, proposal distribution $Q(x, \cdot)$ and new state $y$, the MH test accepts $y$ with probability
\begin{equation}
    \min\{1, \pi(y)Q(y,x) / \pi(x)Q(x,y)\} \label{eq:acc}    
\end{equation}
to guarantee the Markov chain is $\pi$-reversible. Rewriting the simulation \eqref{eq:ld_simu} as a Gaussian proposal distribution
\begin{equation}
    x_{t+\epsilon} \sim Q(x, \cdot) = \mathcal{N}(\cdot; x + \epsilon \nabla \log \pi(x), 2\epsilon I), \label{eq:q_lmc}
\end{equation}
one obtains the \textit{Langevin Monte Carlo} (LMC) sampling algorithm for continuous spaces.



\section{DISCRETE LANGEVIN FRAMEWORK}
Given a finite set $V = \{1, 2, ..., M\}$, a distribution on $V$ is an M-dimensional vector. These vectors form a manifold as the set of $M-1$ dimensional simplex:
\begin{equation}
    \mathcal{P}(V) = \big\{\rho \in \mathbb{R}^M: \sum_{i=1}^M \rho_i = 1, \rho_i \ge 0 \big\}. \label{eq:manifold}    
\end{equation}
For a point $\rho \in \mathcal{P}(V)$, the associated tangent space at $\rho$ \citep{do1992riemannian} is 
\begin{equation}
    T_\rho \mathcal{P}(V) = \big\{\sigma \in \mathbb{R}^M: \sum_{i=1}^M \sigma_i = 0\big\}. \label{eq:tangent}
\end{equation}
We assume the target distribution $\pi \in \mathcal{P}(V)$ is determined by an energy function $f$, such that
\begin{equation}
    \pi_i = \exp(-f_i) / \sum_{k \in V} \exp(-f_k). \label{eq:target}
\end{equation}
Each state $x\in V$ corresponds to a distribution $\rho^0 \in\mathcal{P}(V)$ as a one-hot vector with the $k$-th site equal to $1$.
To find efficient MCMC algorithms for drawing samples $x_t$ from a target $\pi$, we first consider the gradient flow $\rho_t$ that minimizes the KL-divergence $D_\text{KL}(\rho^t || \pi)$.

\subsection{Discrete Wasserstein Gradient Flow}
\label{sec:dfpe}
The gradient flow depends on both the loss function $D_\text{KL}(\rho_t || \pi)$ and the metric in the space. To established the Wasserstein distance in $\mathcal{P}(V)$, we follow \citet{chow2012fokker, chow2017entropy} to rewrite the Wasserstein distance in the language of fluid dynamics via Benamou-Brenier formula \citep{benamou2000computational}:
\begin{equation}
    \footnotesize
    W_2^2(\rho^0, \rho^1):= \inf_v \left\{\int_0^1 \langle v^t, v^t\rangle_{\rho^t} dt: \frac{d\rho^t}{dt} = -\nabla\cdot(\rho^tv^t) \right\}, \label{eq:bbf}
\end{equation}
where $v \in \mathbb{R}^d \rightarrow \mathbb{R}^d$ is a vector field on $\mathbb{R}^d$, and $\langle v, v \rangle_\rho = \frac{1}{2} \int \langle v(x), v(x) \rangle \rho(x) dx$ is the total kinetic energy.

On $\mathcal{P}(V)$, we have a natural generalization of the vector field $v: V \rightarrow \mathbb{R}^M$, where $v_i = (v_{ij})_{j=1}^M$ characterize the amount of the transportation from node $i$ to node $j$. For the divergence and inner product, instead of using the canonical form, we first introduce the conductance 
\begin{equation}
    c_{ij}(\rho) = c_{ji}(\rho) \ge 0
\end{equation} 
between two nodes $i, j \in V$, depending on the current distribution $\rho$, to characterize the conductivity. Then we define the divergence of a vector field $v$ as:
\begin{equation}
    \text{div}_\rho(v) := - \Big(\sum_{i\neq j} c_{ij}(\rho)v_{ji}\Big)_{j=1}^M \in T_\rho \mathcal{P}(V),
\end{equation}
and inner product between two vector fields $u, v$ as:
\begin{equation}
    \langle u, v \rangle_\rho := \frac{1}{2} \sum_{i,j} c_{ij}(\rho) u_{ij} v_{ij}. \label{eq:flow_inner_product}
\end{equation}
Such a divergence $\text{div}_\rho(\cdot)$ and inner product $\langle \cdot, \cdot \rangle_\rho$ induce a 2-Wasserstein distance $W_2(\rho, \nu)$ via the Benamou-Brenier formula in \eqref{eq:bbf} and make the manifold a Riemannian manifold $\mathcal{P}_2(V)$ \citep{chow2012fokker}. 
\begin{equation}
    W_2^2(\rho^0, \rho^1):= \inf_v \left\{\int_0^1 \langle v^t, v^t\rangle_{\rho^t} dt: \frac{d\rho^t}{dt} = \text{div}_{\rho^t}(v^t)\right\}. \label{eq:dbbf}
\end{equation}
Now, we can characterize the Wasserstein gradient flow in Theorem \ref{thm:main}; see Appendix \ref{app:proof} for a complete proof.
\begin{theorem}
\label{thm:main}
On $\mathcal{P}_2(V)$, the gradient flow that minimizes the KL-divergence $D_\text{KL}(\rho^t||\pi)$ is:
\begin{equation}
    \frac{d\rho^t}{dt} = \Big(\sum_{i \neq j} c_{ij}(\rho^t)(f_i + \log \rho^t_i - f_j - \log \rho^t_j)\Big)_{j=1}^M. \label{eq:dfpe}
\end{equation}
\end{theorem}

\subsection{Discrete Langevin Dynamics}
\label{sec:dld}
Equation \eqref{eq:dfpe} gives a discrete analogue of the Wasserstein gradient flow \eqref{eq:fke}. Naturally, the discrete analogue of the Langevin dynamics should be a Markov jump process, as a particle level realization of \eqref{eq:dfpe}. Denote such a Markov jump process as $X^t$, with rate matrix $Q^t$. Then the Kolmogorov forward equation that characterizes the distribution $\rho^t$ for $X^t$ is given by:
\begin{equation}
    \frac{d\rho^t}{dt} = \rho^t Q^t = \Big(\sum_i \rho^t_i Q^t_{ij} \Big)_{j=1}^M. \label{eq:ctmc}
\end{equation}
When $c_{ij}(\rho) = c_{ij}$ is a constant, computing the rate $Q^t$ via \eqref{eq:dfpe} and \eqref{eq:ctmc} requires knowledge of the current distribution $\rho^t$, which is typically intractable. 
However, we find that, inspired by physics, a proper choice of $c_{ij}(\rho)$ as the \textit{conductance} in nonequilibrium chemical reactions \citep{qian2005thermodynamics} can avoid computing $\rho^t$. In particular, set
\begin{equation}
    c_{ij}(\rho) = w_{ij}\frac{g(\pi_j / \pi_i)\rho_i - g(\pi_i/\pi_j) \rho_j}{f_i + \log\rho_i - f_j - \log \rho_j},
\end{equation}
where $w_{ij}$ satisfying $w_{ij} = w_{ji} \in \mathbb{R}$ is an inherent variability between $i$ and $j$, independent of $\pi$ and $\rho$; and $g(\cdot)$ is the locally balanced (LB) function satisfying $g(a) = a g(\frac{1}{a})$
broadly used in recent LB samplers \citep{zanella2020informed}; a more detailed derivation of $c_{ij}(\rho)$ is given in Appendix \ref{app:conductance}.
Such a $c_{ij}(\rho)$ can significantly simplify \eqref{eq:dfpe} to
\begin{equation}
    \frac{d\rho^t}{dt} = \left(\sum_{i\neq j} w_{ij}\Big[\rho^t_i g\left(\frac{\pi_j}{\pi_i}\right) - \rho^t_j g\left(\frac{\pi_i}{\pi_j}\right)\Big] \right)_{j=1}^M. \label{eq:dfpe_simp}
\end{equation}
In this case, we can set the rate matrix $Q^t$ in \eqref{eq:ctmc} as a tractable, time homogeneous matrix $Q$ such that:
\begin{equation}
    Q_{ij} = \left\{
    \begin{array}{cc}
     w_{ij} g\left(\frac{\pi_j}{\pi_i}\right),    & i \neq j  \\
     - \sum_{k\neq i} w_{ik} g\left(\frac{\pi_k}{\pi_i}\right),   & i = j
    \end{array}
    \right. .\label{eq:Q_simp}
\end{equation}
Accordingly, the Markov jump process $X^t$ associated with the Wasserstein gradient flow \eqref{eq:dfpe} can be characterized as a differential equation with respect to the transition probabilities:
\begin{equation}
    \frac{d}{dh}\mathbb{P}(X^{t+h} = j | X^t = i) = w_{ij} g\left(\frac{\pi_j}{\pi_i}\right). \label{eq:dld}
\end{equation}
Since the Markov jump process determined by \eqref{eq:dld} gives a particle realization of the DWGF in \eqref{eq:dfpe_simp}, we refer to it as \textit{discrete Langevin dynamics} (DLD). 


\section{SAMPLING ALGORITHM}
Next, we study how to efficiently simulate the DLD to sample from a discrete space. We consider the state space $V = \mathcal{C}^N = \{1, ..., C\}^N$, where $N$ is the dimension and $\mathcal{C}$ is a code book with elements represented by one-hot vectors. We follow the commonly used assumption~\citep{grathwohl2021oops, sun2021path, zhang2022langevin} that the energy function $f(\cdot)$ in \eqref{eq:target} is differentiable. Since the approximation error of $\langle \nabla f(x), y - x \rangle$ for $f(y) - f(x)$ is repaired by the MH test \eqref{eq:acc}, we ignore such differences in this section. 

Generally, there exist many different choices for simulating the DLD. For example, the Gillespie algorithm \citep{gillespie1977exact} can simulate a continuous-time trajectory exactly, but is computationally expensive. In this work, since we are only interested in the target distribution $\pi$, we can focus on more efficient MCMC algorithms.

\subsection{Casting Previous  Samplers as DLD}
We first show that previous locally balanced (LB) samplers are essentially simulating the \textit{discrete Langevin dynamics} (DLD) by choosing particular discretizations. 

\paragraph{Single Jump} Consider a special case $V = \{0, 1\}^N$. Denote $x, y \in V$ as the current and the next state, LB-1 \citep{zanella2020informed} and GWG \citep{grathwohl2021oops} set
\begin{equation}
    w_{xy} = 1_{\{\sum_{n=1}^N \|x_n - y_n\| = 1\}}; \label{eq:w_hamming}
\end{equation}
that is to say, $w_{xy} = 1$ if and only if there exists an index $n$, such that $y_n \neq x_n$ and for any other $i = 1, ..., n-1, n+1, ..., N$, $x_i = y_i$. Such a weight $w$ restricts the new state $y$ to lie within the 1-Hamming ball of $x$. 
Then, LB-1 and GWG propose the new state $y$ with probability 
\begin{equation}
    q(x, y) \propto w_{xy} g\left(\frac{\pi(y)}{\pi(x)}\right). \label{eq:prop_q}
\end{equation}
Such a categorical distribution is exactly the first transition probability for $X^t$ satisfying DLD \eqref{eq:dld}. Specifically, we denote the jumping time of $X^t$ as $J_1, ..., J_m, ...$, and assume $X^t$ jumps to $y$ from $x$ at time $J_m$, then we have
\begin{equation}
    \mathbb{P}(X^{J_m} = y| X^{J_{m-1}} = x) \propto w_{xy} g\left(\frac{\pi(y)}{\pi(x)}\right). \label{eq:first_jump}
\end{equation}
A more detailed derivation for \eqref{eq:first_jump} is given in Appendix \ref{app:single}. Hence, one can claim that LB-1 and GWG propose the new state by simulating the first jump of DLD. 

\paragraph{Multiple Jumps} Denote the current state as $\sigma^0 \in V$ and the path length as $L$. PAS \citep{sun2021path, sun2022optimal} propose a new state $\sigma^L \in V$ along the auxiliary path $\sigma$. Specifically, they sequentially propose $\sigma^{l-1} = x, \sigma^l = y$ from
\begin{equation}
    q(x, y) \propto w_{xy} g\left(\frac{\pi(y)}{\pi(x)}\right).
\end{equation}
Similar to the analysis of single jump samplers, for process $X^t$ with the first jump after $J_{l-1}$ occurs at time $J_l$, we have:
\begin{equation}
    \mathbb{P}(X^{J_l} = y| X^{J_{l-1}} = x) \propto w_{xy} g\Big(\frac{\pi(y)}{\pi(x)}\Big).
\end{equation}
Hence, one can claim that PAS propose a new state by simulating the first $L$ jumps of DLD. 

\paragraph{Parallel Jumps}
Denote the current state as $x \in V$. Recent work \citep{zhang2022langevin, rhodes2022enhanced} has generalized the Gaussian proposal in LMC \eqref{eq:q_lmc} to discrete spaces by restricting the proposal distribution to discrete points:
\begin{talign}
    q(x, y) &\propto \prod_{n=1}^N \exp(- r_n(x, y_n)) \label{eq:euler} \\
    r_n(x, y_n) &= \langle y_n - x_n, \frac{\partial}{\partial x_n} f(x)\rangle + \frac{\|x_n - y_n\|}{2\alpha} \label{eq:euler_2}
\end{talign}
where each dimension for $y=(y_1, ..., y_N)$ are sampled independently, which allows convenient parallel implementation. The proposal \eqref{eq:euler} can be seen as setting 
\begin{equation}
    w_{xy} = 1, \quad h = \exp\Big(-\frac{1}{2\alpha}\Big)
\end{equation}
and using the forward Euler's method to approximate \eqref{eq:dld} with simulation time $h$:
\begin{equation}
\mathbb{P}(X_n^{t+h}=y_n| X^t=x) \propto h w_{xy} g\Big(\frac{\pi(x_{\backslash n}, y_n)}{\pi(x_{\backslash n}, x_n)}\Big).
\end{equation}
This value is correct when $y_n \neq x_n$. However, such a Gaussian proposal, copied from a continuous diffusion process, does not use the correct diagonal rate for a discrete jump process. In particular, \eqref{eq:euler} has
\begin{equation}
    \mathbb{P}(X_n^{t+h} = x_n|X^t = x) \propto 1,
\end{equation}
which corresponds to a rate $r_n(x, x_n) = 0$ on the diagonal in the rate matrix. However, in a jump process, the correct rate on the diagonal should be $r_n(x, x_n) = - \sum_{y_n\neq x_n} r(x, y_n)$ the negative summation of the off-diagonal entries. Such a mismatch 
reduces the quality of the proposal distribution; see Appendix \ref{app:dlmcf} for a more detailed discussion. Also, one can effectively improve the sampling efficiency via correcting the rate in diagonal; see more results in Appendix \ref{app:experiment}.



\subsection{Discrete Langevin Monte Carlo}
\label{sec:dlmc}
The framework induced by the \textit{discrete Wasserstein gradient flow} (DWGF) in \eqref{eq:dfpe_simp} provides a more principled way to design gradient based MCMC algorithms by estimating the transition probability matrix. In particular, since the gradient flow is a Markov jump process, we have the closed form for the transition:
\begin{equation}
    \rho^{t+h} = \rho^t P^h, \quad P^h = \exp(Q h),
\end{equation}
for $Q \in \mathbb{R}^{C^N \times C^N}$ in \eqref{eq:Q_simp}. Of course, it is impractical to directly calculate the matrix exponential $\exp(Q s)$ for a large rate matrix $Q$. Instead, by factorizing the jump process $X^t=(X^t_1, X^t_2, ..., X^t_N) = (x^t_1, x^t_2, ..., x^t_N)$, one can simulate each sub-processes $X^t_n$ with initial value $x^t_n$, independently.
In this case, the distribution $\rho^t_n$ for $X^t_n$ has the following closed form expression for the transition:
\begin{equation}
    \rho_n^{t+h} = \rho_n^t P_n^h(x^t), \quad P_n^h(x^t) = \exp(Q_n(x^t) h), \label{eq:P_transition}
\end{equation}
where the rate matrix $Q_n(x^t) \in \mathbb{R}^{C \times C}$ depends on the current state $x^t$.
In particular, for index $i \neq j \in \mathcal{C}$, the rate matrix $Q_n(x^t)$ satisfies 
\begin{equation}
    Q_n(x^t)(i, j) = w_{ij} g\left(\pi(x^t_{\backslash n}, j) / \pi(x^t_{\backslash n}, i) \right) .
\end{equation}
For simplicity, we will drop the $x^t$ and only use $P_n^h(i,j)$, $Q_n(i, j)$ when this does not cause ambiguity.
For the binary case $C = 2$, denoting $\alpha = Q_n(1, 2)$ and $\beta = Q_n(2, 1)$, then the transition matrix $P_n^h$ has a closed form expression \eqref{eq:exp_2}:
\begin{equation}
P^h_n = \left(\begin{array}{ll}
\frac{\beta}{\alpha+\beta}+\frac{\alpha}{\alpha+\beta} e^{-(\alpha+\beta) h} & \frac{\alpha}{\alpha+\beta}-\frac{\alpha}{\alpha+\beta} e^{-(\alpha+\beta) h} \\
\frac{\beta}{\alpha+\beta}-\frac{\beta}{\alpha+\beta} e^{-(\alpha+\beta) h} & \frac{\alpha}{\alpha+\beta}+\frac{\beta}{\alpha+\beta} e^{-(\alpha+\beta) h}
\end{array}\right). \label{eq:exp_2}
\end{equation}
One can sample $y_n$ from the $x^t_n$-th row of $P_n^g$, a categorical distribution, for $n=1, ..., N$ in parallel. Hence, the new state $y = (y_1, ..., y_N)$ can be efficiently obtained.

For the categorical case $C > 2$, we do not have a simple closed form expression of $P_n^h$ for all $C$. Instead, we generalize the expression in \eqref{eq:exp_2}. Denote 
\begin{equation}
    \textstyle
    \nu_n(x^t)(j) = \pi(x^t_{\backslash n}, j) / \sum_{i=1}^C \pi(x^t_{\backslash n}, i) \label{eq:stationary}
\end{equation}
as the stationary distribution induced by $Q_n(x^t)$. For simplicity, we drop $x^t$ and only use $\nu_n(j)$ when this does not cause ambiguity. We approximate the transition as:
\begin{equation}
    \tilde{P}_n^h(i,j) = \left\{  
    \begin{array}{cc}
    \nu_n(i) + \sum_{k\neq i} \nu_n(k) e^{-h \frac{Q_n(i, k)}{\nu_n(k)}},     & i =j \\
    \nu_n(j) - \nu_n(j) e^{- h\frac{Q_n(i, j)}{\nu_n(j)}},   & i \neq j
    \end{array}
    \right. . \label{eq:P_interpolation}
\end{equation}
Such an approximation is consistent with the special case $C=2$ in \eqref{eq:exp_2}, and satisfies the boundary conditions $\tilde{P}_n^0 = P_n^0$, $\tilde{P}_n^\infty = P_n^\infty$, $\frac{d}{dh}\tilde{P}_n^h|_{h=0} = \frac{d}{dh}P_n^h|_{h=0}$ for arbitrary $C$. Hence, Equation \eqref{eq:P_interpolation} provides a better approximation than the forward Euler's method. In practice, we find \eqref{eq:P_interpolation} does not lose much proposal quality compared to calculating the matrix exponential for $P_n^h$ in \eqref{eq:P_transition}. On the other hand, we only need to compute the $x_n^t$-row in $\tilde{P}_n^h$, with computational cost is $O(C)$. This is much more efficient than a generic numerical approximation of the matrix exponential with cost $O(C^3)$ \citep{al2010new}.

\begin{algorithm}[t]
  \KwInput{current state $x^t$, step time $h$, target $\pi$}
  \KwOutput{new state $x^{t+h}$}
  \For(\tcp*[f]{Run in parallel}){n=1, ..., N}{
  Calculate $x^t_n$-row of $\tilde{P}_n^h(x^t)$ in \eqref{eq:P_interpolation} \\
  Sample $y_n \propto \tilde{P}_n^h(x^t)(x^t_n ,y_n)$
  }
  Compute $A = \min\{1, \frac{\pi(y) \prod_{n=1}^N \tilde{P}^h_n(y)(y_n, x_n^t)}{\pi(x) \prod_{n=1}^N \tilde{P}^h_n(x^t)(x_n^t, y_n)}\}$ \\
  \lIfElse{$\text{rand}(0,1) < A$}{$x^{t+h}=y$}{$x^{t+h}=x^t$}
\caption{DLMC MH step}
\label{alg:main}
\end{algorithm}

In a concurrent work \citep{sun2023anyscale}, the ``globally balanced'' phenomenon is observed, where using $g(a) = a^\alpha$ with $\alpha \in (0.5, 1]$ can have better performance than that with $\alpha = 0.5$ in some distributions. Such a phenomenon occurs when the target distribution is nearly factorized, for example Bernoulli distribution in section \ref{sec:exp_bernoulli}. In DLMC, the selection of $\alpha$ is implicitly done when tuning the simulation time $h$. Specifically, when $h$ is small, the transition probability \eqref{eq:P_interpolation} is equivalent with using forward Euler's method  with $g(a) = a^\frac{1}{2}$. When $h$ is large, the transition probability degenerates to the stationary distribution $\nu$ in \eqref{eq:stationary}, whose value can be obtained by using $g(a) = a$. This explains the ``globally balanced'' phenomenon is originated from a large simulation time of DLD.

Combining the DLD and the discretization via \eqref{eq:P_interpolation}, we obtain the \textit{Discrete Langevin Monte Carlo} (DLMC) algorithm.  
The discritization in \eqref{eq:P_interpolation} not only provides a factorized proposal distribution for parallel computing, but also gives a time-uniform slicing of DLD. By contrast, PAS \citep{sun2021path, sun2022optimal} only flips a fixed number of sites in each MH step, thereby has a simulation time that depends on the current state. Specifically, PAS has a shorter simulation time at states with a larger jump rate, and longer simulation time at states with a smaller jump rate. Consequently, PAS realizes a non-uniform time slicing of DLD, which leads to more proposal rejections in comparison to DLMC; see Appendix \ref{app:dlmc} for more details.
Pseudo code for an MH step of DLMC is given in Algorithm \ref{alg:main}.

\section{RELATED WORK}
Gradient based MCMC algorithms that simulate Langevin dynamics \citep{rossky1978brownian, girolami2011riemann, welling2011bayesian} or Hamiltonian dynamics \citep{duane1987hybrid, neal2011mcmc, hoffman2014no}, can substantially improve sampling efficiency in both theory and practice. The seminal work of \citet{jordan1998variational} and \citet{otto2001geometry} shows that the Langevin dynamics simulates the gradient flow on the 2-Wasserstein space $\Pcal_2(\mathbb{R}^D)$ \citep{villani2009optimal}. Subsequent work has directly studied the Wasserstein gradient flow \citep{mokrov2021large} or extended the result to Hamiltonian dynamics \citep{ambrosio2008hamiltonian, liu2019understanding, chow2020wasserstein} and particle variational inference \citep{chen2018unified, liu2019understanding}. 
By contrast, the corresponding theory for sampling algorithms in discrete spaces is less well understood. \citet{mielke2011gradient, maas2011gradient, chow2012fokker} introduce 2-Wasserstein distances on finite graphs via the Benamou-Brenier formula \citep{benamou2000computational}. However, these works do not investigate Langevin dynamics or sampling algorithms in discrete space.

A number of samplers for discrete spaces construct invertible mappings between discrete and continuous spaces via auxiliary variables, uniform dequantization, or VAE flow \citep{zhang2012continuous, pakman2013auxiliary, nishimura2017discontinuous, han2020stein, jaini2021sampling}. Such methods work in some scenarios, but a key challenge is that embedding the discrete space in a continuous space can destroy the inherent discrete structure, resulting in irregular target distributions in the continuous space such that compromises performance in high dimensional discrete spaces \citep{grathwohl2021oops}.

Another group of methods work directly on discrete spaces.
\citet{dai2020learning, titsias2017hamming} augment the discrete space with an auxiliary variable, but still rely on slow Gibbs sampling for improvement.
\citet{zanella2020informed} introduces an informed proposal for discrete spaces, and proves that a family of locally balanced (LB) functions is asymptotically optimal. Following this work, \citet{grathwohl2021oops, sun2021path, zhang2012continuous, rhodes2022enhanced, sun2022optimal, sun2023anyscale} propose various LB samplers. Despite these LB samplers substantially improving sampling efficiency in discrete spaces by mimicking LMC, their lack of a principled connection to the \textit{discrete Langevin dynamics} (DLD) results in sub-optimal proposal distribution designs.  We note that special cases of DLD \eqref{eq:dld} have been mentioned in previous work \citep{sohl2009minimum, power2019accelerated} but without realizing the connection to gradient flow.

\section{SAMPLING FROM CLASSICAL EBMS}
\label{sec:sampling}
\vspace{-1mm}
\subsection{Settings}
\paragraph{Models}
We demonstrate the advantage of DLMC in sampling tasks on four classical models: the Bernoulli model (Bernoulli), Ising model (Ising) \citep{ising1924beitrag}, factorial hidden Markov model (FHMM) \citep{ghahramani1995factorial}, and the restricted Boltzmann machine (RBM) \citep{mcclelland1987parallel}. For each model, we use a binary version $C = 2$ with high or low temperature, a 4-category version $C = 4$, and an 8-category version $C = 8$. Compared to the low temperature model, the high temperature model is smoother and has larger entropy. Here we only report the results on high temperature version and 8-category version. More description of the models and additional results are given in Appendix \ref{app:experiment}.

\paragraph{Baselines}
We consider the LB samplers GWG \citep{grathwohl2021oops}, PAS \citep{sun2022optimal}, and DMALA \citep{zhang2022langevin}. Note that, for PAS, we follow the implementation in \citet{sun2022optimal}, which is computationally more efficient compared to the PAS in the original paper \citep{sun2021path}. Also, NCG \citep{rhodes2022enhanced} is equivalent to DMALA, so we do not report results for NCG. For these LB samplers, we consider two commonly used weight functions $g(a) = \sqrt{a}$ and $g(a) =\frac{a}{a+1}$. Also, we select the optimal hyperparameters by tuning the average acceptance rate to $0.574$ following the result \citep{sun2022optimal}.
In particular, we tune
$U$, how many sites to flip per MH step for PAS, $\alpha$, the step size for DMALA, and $h$, the simulation time for DLMC.  Although the optimality for $0.574$ is only proved for PAS, we find it robustly produce good results for DMALA and DLMC, so we still use this technique.

We also compare with classical discrete samplers: random walk Metropolis (RWM), the Hamming Ball sampler (HB) \citep{titsias2017hamming}, and block Gibbs (BG). Following \citet{grathwohl2021oops}, we use a block size of 10 and hamming distance 1 for HB, and a block size of 2 for BG. Silimar to PAS, we also select an optimal $U$, how many sites to flip per MH step, for RWM by setting the average acceptance rate to $0.234$ to achieve the otpimal efficiency \citep{sun2022optimal}.

\paragraph{Metrics}
We use effective sample size (ESS) to evaluate each sampler \citep{lenth2001some}. To reduce the effects of implementation, we report ESS normalized by the number of energy evaluations, and the running time. For methods requires gradients, we count each gradient backpropagation as one call of the energy function as they have the similar computational cost. The former measure ignores the computational cost for sampling a new state from the proposal distribution and focuses on proposal quality. The latter measure reflects proposal efficiency. For each setting and sampler, we run 100 chains for 100,000 steps, with 50,000 burn-in steps to ensure the chain mixes.

\subsection{Results}
\label{sec:some_results}
\begin{figure*}
    \centering
    \includegraphics[width=0.945 \textwidth]{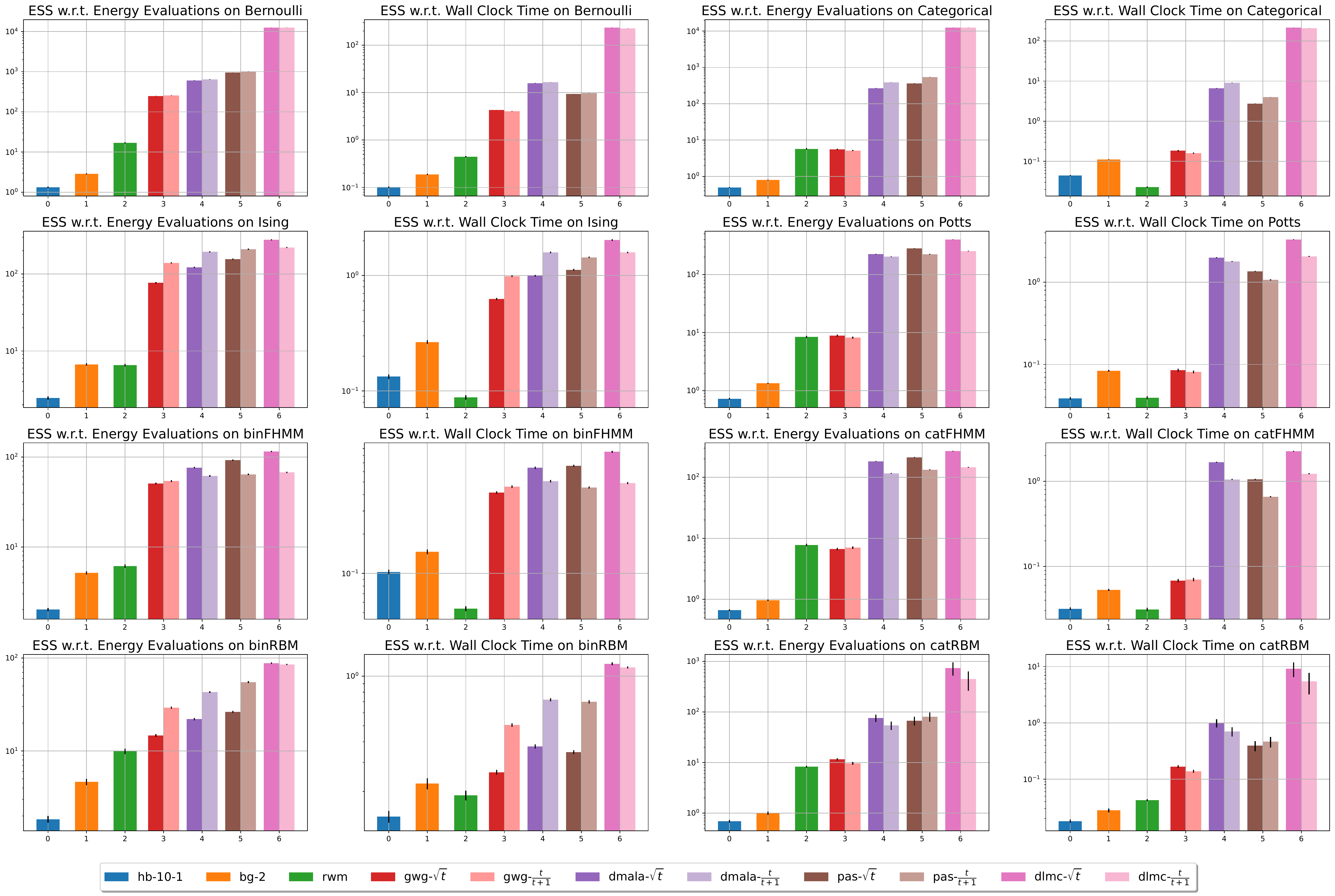}
    \caption{Effective Sample Size ($\uparrow$) on Various Distributions in log scale}
    \label{fig:result_sampling}
\end{figure*}

\paragraph{Bernoulli and Categorical}
\label{sec:exp_bernoulli}
The Bernoulli distribution is the simplest distribution in a discrete space, consisting of independent binary random variables. The categorical distribution is a simple generalization to categorical random variables. For $x \in \mathcal{C}^N$, the energy function is:
\begin{equation}
    f(x) = \sum_{n=1}^N \langle x_n, \theta_n\rangle
\end{equation}
We report the results in the first row of \figref{fig:result_sampling}. We can see that the DLMC significantly outperforms all the other samplers. On Bernoulli and categorical distributions, DLMC has ESS with respect to energy evaluation larger than $10^4$. Since each step of DLMC requires 4 evaluations of energy function, it basically means the 50k samples collected by DLMC are independent. The reason is that DLMC does not lose accuracy in factorizing. With a simulation time large enough, the proposal distribution in \eqref{eq:P_interpolation} is exactly the target distribution. 

Another interesting observation is that, compared to PAS, DMALA has a smaller ESS with respect to the number of energy evaluations, but a larger ESS with respect to the running time. The reason is that DMALA does not correctly simulate the WGF, which reduces its proposal quality. PAS generates the new state by constructing an auxiliary path sequentially, where the lack of parallelism reduces efficiency.

\paragraph{Ising and Potts}
The Ising model \citep{ising1924beitrag} is a mathematical model of ferromagnetism in statistical mechanics. It consists of binary random variables arranged in a lattice graph $G=(V, E)$ and allows each node to interact with its neighbors. The Potts model \citep{potts1952some} is a generalization of the Ising model where the random variables are categorical. The energy function is:
\begin{equation}
    f(x) = - \sum_{n=1}^N \langle x_n, \theta_n\rangle - \lambda \sum_{(i, j) \in E} \delta(x_i, x_j)
\end{equation}
We report the results in the second row of \figref{fig:result_sampling}. We can see that the advantage of DLMC narrows compared to Bernoulli model, as Ising and Potts models are not factorized, but the gap is still significant. Also, one can notice an interesting phenomenon that all LB samplers, except for DLMC, have large ESS with $g(a) = \frac{a}{a+1}$ in binary models and with $g(a) = \sqrt{a}$ in categorical models.

\paragraph{FHMM}
The Factorial Hidden Markov Model \citep{ghahramani1995factorial} uses latent variables to characterize time series data. In particular, it assumes the continuous data $y \in \mathbb{R}^L$ is generated by hidden state $x \in \mathcal{C}^{L\times K}$. When $C = 2$, we call it a binary FHMM (binFHMM) and when $C > 2$, we call it a categorical FHMM (catFHMM). The probability function is:
\begin{align}
    p(x) &= p(x_1) \prod_{l=2}^L p(x_t|x_{t-1}) \\
    p(y|x) &= \prod_{l=1}^L \mathcal{N}(y_t; \sum_{k=1}^K \langle W_k, x_{l, k}\rangle + b; \sigma^2)
\end{align}
We report the results in the third row of \figref{fig:result_sampling}.
Similar to the Ising model, we can see that all LB samplers demonstrate good efficiency and DLMC still leads the performance. In FHMM, we can see that using $g(a) = \sqrt{a}$ is more efficient than using $g(a) = \frac{a}{a+1}$ across all LB samplers. One possible reason is that using $g(a) = \sqrt{a}$ is more likely to jump to high probability states that makes the sampling more efficient on smooth target distributions, but less robust on nonsmooth target distributions \citep{livingstone2019barker}.

\paragraph{RBM}
The Restricted Boltzmann Machine \citep{smolensky1986information} is an unnormalized latent variable model, with a visible random variable $v \in \mathcal{C}^N$ and a hidden random variable $h \in \{0, 1\}^M$. When $C = 2$, we call it a binary RBM (binRBM) and when $C > 2$, we call it a categorical RBM (catRBM). The energy function is:
\begin{align}
    f(v) = \sum_{h} \big[ & - \sum_{n=1}^N \langle v_n, \theta_n\rangle - \sum_{m=1}^M \beta_m h_m \notag \\
    & -\sum_{n, m} \langle h_m \theta_{m, n}, v_n \rangle \big]
\end{align}
Unlike the previous three models, where the parameters are hand designed, we train binRBM on MNIST \citep{lecun1998gradient} and catRBM on Fashion-MNIST \citep{xiao2017/online} using contrastive divergence \citep{hinton2002training}. The learned RBMs have stronger multi-modality compared to previous models and are harder to sample from. We report the results in the fourth row of \figref{fig:result_sampling}. We can see that DLMC is significantly more efficient than all the other samplers with respect to both number of energy evaluations and the running time. Also, we can see that in binRBM, although $g(a) = \frac{a}{a+1}$ is significantly more efficient in other LB samplers, DLMC still has larger ESS using $g(a) = \sqrt{a}$.


\section{LEARNING DEEP EBMS}
\begin{table*}[t]
    \centering
\caption{Evaluation of effectiveness on learning binary EBMs. \label{tab:ebm_bin}}
\resizebox{1.0\textwidth}{!}{%
    \begin{tabular}{c|c|cccccccccc}
\toprule
    \multirow{2}{*}{Data Type} & \multirow{2}{*}{Dataset} & VAE & VAE & RBM & DBN & EBM & EBM & EBM & EBM & EBM \\
    & & (MLP) & (Conv) & & & (GWG) & (Gibbs) & (PAS) & (DMALA) & (DLMC) \\
\hline
    \multirow{2}{*}{Binary} & Static MNIST & -86.05 & -82.41 & -86.39 & -85.67 & -80.01 & -117.17 & -79.58 &  -79.46 & {\bf -79.13}\\
    & Dynamic MNIST & -82.42 & -80.40 & $-$ & $-$ & -80.51 & -121.19 & -79.59 & -79.54 & {\bf -78.84} \\ 
    \multirow{2}{*}{log-likelihood $\uparrow$ } & Omniglot & -103.52 & 97.65 & -100.47 & -100.78 & -94.72 & -142.06 & {\bf -90.75} & -91.11 & -90.84 \\
    & Caltech Silhouettes & -112.08 & -106.35 & $-$ & $-$ & -96.20 & 163.50 & -84.56 & -87.82 & {\bf -77.04} \\
\bottomrule
    \end{tabular}
}%
\end{table*}
\begin{figure*}[t]
    \centering
\begin{tabular}{@{}c@{\hskip 1mm}c@{\hskip 1mm}c@{\hskip 1mm}c@{\hskip 1mm}c}
    \includegraphics[width=0.2\textwidth]{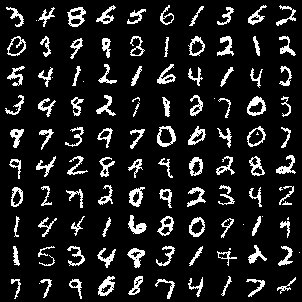} & 
    \includegraphics[width=0.2\textwidth]{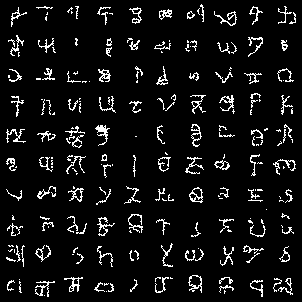} &
    \includegraphics[width=0.2\textwidth]{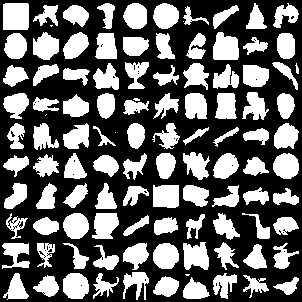} &
    \includegraphics[width=0.15\textwidth]{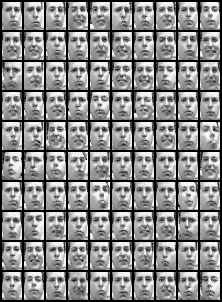} &
    \includegraphics[width=0.2\textwidth]{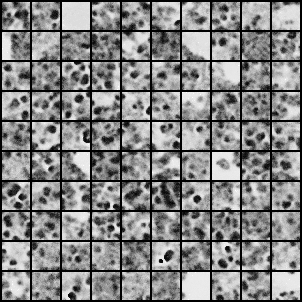} \\
    Binary MNIST & Omniglot & Caltech & Frey Faces & Histopathology \\
\end{tabular}
    \caption{Samples from learned deep EBMs using proposed LBJ sampler. \label{fig:deep_samples}}
\end{figure*}
Deep energy-based models (EBMs) have gained increasing popularity. Recent advances including tempered Langevin samplers \citep{nijkamp2020anatomy}, large persistent chains \citep{du2019implicit}, and amortized sampling~\citep{dai2019exponential,dai2020learning}, which have enabled deep EBMs to become a competitive approach for generative modeling \citep{song2020score, sun2021towards, xie2021mars, bakhtin2021residual}.
However, learning an EBM is challenging. Given data sampled from a true distribution $\pi$, we maximize the likelihood of the target distribution $\pi_\theta(x) \propto e^{-f_\theta(x)}$ parameterized by $\theta$. The gradient of the likelihood is:
\begin{equation}
    \nabla_\theta \log \pi_\theta(x) = \mathbb{E}_\pi [\nabla_\theta f_\theta(x)] - \mathbb{E}_{\pi_\theta} [\nabla_\theta f_\theta(x)]
\end{equation}
The first expectation can be estimated using the data from the true distribution. The second expectation requires samples from the current model, which is are typically obtained via MCMC. The speed of EBM training is determined by how fast the MCMC algorithm can obtain a good estimate of the second expectation. Following \citet{grathwohl2021oops} and \citet{sun2021path}, we train deep EBMs parameterized by residual networks \citep{he2016deep} on binary and grayscale image datasets using PCD \citep{tieleman2008training} with a replay buffer \citep{du2019implicit}. The grayscale images were treated as 1-of-256 categorical data.

\begin{table}[ht]
    \centering
\caption{Evaluation on learning categorical EBMs. \label{tab:ebm_cat}}
\resizebox{0.49\textwidth}{!}{%
    \begin{tabular}{c|c|cccccc}
\toprule
    \multirow{2}{*}{Data Type} & \multirow{2}{*}{Dataset} & VAE & VAE & EBM & EBM & EBM \\
    & & (MLP) & (Conv) & (GWG) & (PAS) & (DLMC) \\
\hline
Categorical & Frey Faces & 4.61 & 4.49 & 4.65  & 4.74 & {\bf 4.33} \\
(bits/dim $\downarrow$) & Histopathology & 5.82 & 5.59 & 5.08 & 5.1 & {\bf 4.91} \\
\bottomrule
    \end{tabular}
}%
\end{table}
We present the test-set likelihoods in Table \ref{tab:ebm_bin} and Table \ref{tab:ebm_cat}. Likelihoods are estimated using annealed importance sampling \citep{neal2001annealed}. 
Since the quality of the learned EBMs will be similar as long as the sampler is good enough with certain steps per model update, we measure the efficiency of samplers by the minimum number of MCMC steps needed to chain a decent EBM.
We compare the performance of DLMC to Variational Autoencoder \citep{kingma2013auto}, an RBM, a deep belief network (DBN) \citep{hinton2009deep} and EBMs trained by Gibbs, GWG \citep{grathwohl2021oops}, PAS \citep{sun2021path, sun2022optimal}, and DMALA \citep{zhang2022langevin}. We use weight function $g(t) = \sqrt{t}$ for all LB samplers. On all datasets, the DLMC samplers enable deep EBMs to become competitive on high dimensional discrete data. We also present long-run samples from the EBMs trained by DLMC in Figure \ref{fig:deep_samples}.

\section{DISCUSSION}
\label{sec:discussion}
We have described the \textit{discrete Langevin dynamics} (DLD) and showed that it simulates the \textit{discrete Wasserstein gradient flow} (DWGF) to minimize the KL-divergence to a target distribution. Such a view provides an unified framework to design gradient based samplers for discrete spaces. Based on this perspective, we proposed a new algorithm, \textit{discrete Langevin Monte Carlo} (DLMC), that improves the efficiency of existing locally balanced samplers in both sampling and learning tasks across various discrete distributions.
Despite the success of  DLMC presented in this work, there remain several interesting problems to investigate. 

The DWGF and DLD are determined by  topological structure via $w_{ij}$ and $g(\cdot)$ \eqref{eq:conductance}. For $w_{ij}$, in complicated discrete spaces like Hamiltonian cycles, one can resort to powerful heuristics to construct shortcuts between a current state $i$ and a new state $j$. For the weight function, we empirically evaluated the most commonly used alternatives $g(t) = \sqrt{t}$ and $g(t)=\frac{t}{t+1}$ in Sec. \ref{sec:some_results} and found that each has its own advantages on different models. Although \citet{sansone2021lsb} has some initial attempts to learn $g(\cdot)$ as a linear combination of 4 commonly used weight functions, a principled understanding of the weight function is still missing. In future work, we will investigate whether DWGF can be used as a tool to analyze the choice of $g(\cdot)$.

After obtaining DLD, one can use any kind of discretization to obtain a sampling algorithm. Besides the DLMC and previous LB samplers, there are many other choices. For example, we can use forward Euler's method to simulate DLD, which we call DLMCf. DLMCf can be seen as DMALA with a corrected diagonal rate. In Appendix \ref{app:experiment}, we show that DLMCf has comparable performance with DLMC on several models, and is substantially more efficient than DMALA. Considering fruitful numerical methods like Heun's method and Runge-Kutta method can outperforms forward Euler's method in many scenarios, we believe there is ample room to improve the discretization.

Overall, DWGF and DLD provide a new framework for sampling algorithms in discrete spaces. We believe this is a milestone for sampling in discrete spaces and expect further development upon this framework.

\subsubsection*{Acknowledgements}
We thank four anonymous
reviewers for their helpful comments to improve the manuscript.
Dale Schuurmans gratefully acknowledges support from a CIFAR Canada AI Chair, NSERC and Amii.

\bibliography{reference}
\bibliographystyle{icml}

\onecolumn
\appendix
\addcontentsline{toc}{section}{Appendices}
\section*{Appendices}

\section{COMPLETE PROOFS}
\label{app:proof}
In order to formally prove theorem \ref{thm:main}, we need the following two lemmas.
\begin{lemma}
\label{lemma:potential}
For any $\rho \in \Pcal(V)$ and any vector field $u$, the minimizer $v^* = \argmin_v \langle v, v \rangle$, subject to $\div_\rho(v) = \div_\rho(u)$, is a potential field $\nabla \Phi$. That is, there exists a function $\Phi: V \rightarrow \mathbb{R}$, such that $v^*_{ij} = \nabla \Phi_{ij} = (\Phi_i - \Phi_j) 1_{\{(i, j) \in E\}}$.
\end{lemma}
We note that a potential field is invariant up to a constant shift, meaning that if $\Phi$ is a potential function and $\Phi' = \Phi + c = (\Phi_i + c)_{i=1}^M$, then $\nabla \Phi' = \nabla \Phi$. 
Hence, we consider an equivalence class $[\Phi] = \{ \Phi' \in \mathbb{R}^M: \exists c \in \mathbb{R}, \Phi' = \Phi + c\}$ and denote $P^M = \{[\Phi]: \Phi \in \mathbb{R}^M\}$
\begin{lemma}
\label{lemma:isomorphism}
For any $\rho \in \Pcal(V)$, the mapping $\zeta([\Phi]) = \div_\rho(\nabla \Phi)$ is a linear isomorphism between the set of equivalence classes $P^M$ and the tangent space $T_\rho \Pcal(V)$.
\end{lemma}
In this case, the isomorphism $\zeta$ induces a metric on the tangent space $T_\rho \Pcal(V)$:
\begin{definition}
\label{def:metric}
For any $\rho \in \Pcal(V)$, we define the inner-product $\langle \cdot, \cdot \rangle_\rho$ on $T_\rho \Pcal(V)$ as follows. Denote $\Phi^\sigma \in \zeta^{-1}(\sigma)$. Then for arbitrary $\sigma^1, \sigma^2 \in T_\rho \Pcal(V)$, define
\begin{equation}
    \langle \sigma^1, \sigma^2 \rangle_\rho = \sum_{i=1}^M \sigma^1_i \Phi^{\sigma^2}_i \label{eq:metric}
\end{equation}
\end{definition}
In the following section, we first prove lemma \ref{lemma:potential} and lemma \ref{lemma:isomorphism}, then we justify definition \ref{def:metric} is well-defined, and finally we prove theorem \ref{thm:main}.

\subsection{Proof for Lemma \ref{lemma:potential}}
Consider $V = \{1, ..., M\}$. Denote $F(G)$ as the set of all vector fields on graph $G$. The divergence operator $\div_\rho$ maps a vector field $v \in F(G)$ to a vector $\sigma$ in the tangent space $T_\rho \Pcal(V)$. It is not hard to see that $\div_\rho$ is a surjection, but not an injection. For a $\sigma \in T_\rho \Pcal(V)$, there are infinite choices of vector field $v$ such that $\div_\rho(v) = \sigma$. Lemma \ref{lemma:potential} tells us that for Wasserstein distance defined in \eqref{eq:dbbf}, we only need to consider vector field $v$ as a potential field.
\begin{proof}
We show that given arbitrary vector field $u$, there exists a potential field $\nabla \Phi$ has the same divergence and minimizes the norm. In particular, let us consider the following optimization problem:
\begin{equation}
    \min_v \langle v, v \rangle, \quad \text{ subject to: } \div_\rho(v) = \div_\rho(u) = \sigma \label{eq:opt}
\end{equation}
We introduce the dual variable $(\lambda_i)_{i=1}^M$ and we have the Lagrangian:
\begin{align}
    L(v, \lambda) 
    &= \frac{1}{2} \sum_{(i, j) \in E} c_{ij} v_{ij}^2 + \sum_{i=1}^M \lambda_i (\sigma_i - \sum_{j \in N(i)} c_{ji} v_{ji} - c_{ij} v_{ij}) \\
    &= \sum_{(i, j) \in E} [(\lambda_i - \lambda_j) + \frac{1}{2} v_{ij}] c_{ij} v_{ij} + \sum_{i=1}^M \lambda_i \sigma_i
\end{align}
Since $u$ is a solution, the optimization problem \eqref{eq:opt} is feasible. Since the inner product $\langle v, v\rangle \ge 0$, the optimization problem \eqref{eq:opt} is bounded. By Slater's condition, the strong duality holds and the Lagrangian is minimized at $(v^*, \lambda^*)$ with a finite value. Hence, we have $v^*_{ij} = \lambda^*_j - \lambda^*_i$. When we let $\Phi_i = \lambda^*_i$, we have $v^* = \nabla \Phi$ is a potential field.
\end{proof}

\subsection{Proof for Lemma \ref{lemma:isomorphism}}
Lemma \ref{lemma:potential} tells us that the minimum vector field to realize a divergence is in the form of a potential field. We can notice that a potential field is invariant up to a constant shifting. That's to say, if $\Phi$ is a potential function and $\Phi' = \Phi + c = (\Phi_i + c)_{i=1}^M$, then $\nabla \Phi' = \nabla \Phi$. 
Hence, we consider a equivalence class $[\Phi] = \{ \Phi' \in \mathbb{R}^M: \exists c \in \mathbb{R}, \Phi' = \Phi + c\}$ and we denote $P^M = \{[\Phi]: \Phi \in \mathbb{R}^M\}$. Then, lemma \ref{lemma:isomorphism} gives an isomorphism between $P^M$ and $T_\rho \Pcal(V)$.
\begin{proof}
We first show $\zeta([\Phi]) = \div_\rho(\nabla \Phi)$ is well-defined. For arbitrary $\Phi^1, \Phi^2 \in [\Phi]$, we have $\nabla \Phi^1 = \nabla \Phi^2$, thereby $\div_\rho(\nabla \Phi^1) = \div_\rho(\nabla \Phi^2)$. It indicates $\zeta$ is well-defined.

Second, we show $\zeta$ is linear. We have
\begin{align}
    \zeta(\alpha [\Phi^1] + \beta [\Phi^2]) 
    &= \zeta([\alpha \Phi^1 + \beta \Phi^2]) \label{eq:equiv_sum} \\
    &= \div_\rho(\nabla (\alpha \Phi^1 + \beta \Phi^2) ) \\
    &= \alpha \div_\rho(\nabla \Phi^1) + \beta \div_\rho(\nabla \Phi^2) \\
    &= \alpha \zeta([\Phi^1]) + \beta \zeta([\Phi^2])
\end{align}
We have \eqref{eq:equiv_sum} holds as
\begin{align}
     \psi \in \alpha [\Phi^1] + \beta [\Phi^2]
    \Longleftrightarrow  & \exists c^1, c^2, \psi = \alpha (\Phi^1 + c^1) + \beta (\Phi^2 + c^2) \\
    \Longleftrightarrow  & \exists c, \psi = \alpha \Phi^1 + \beta \Phi^2 + c \\
    \Longleftrightarrow  & \psi \in [\alpha \Phi^1 + \beta \Phi^2]
\end{align}
Third, we show that $\zeta$ is an injection. By the property shown above, we have
\begin{equation}
    \zeta([\Phi^1]) = \zeta([\Phi^2]) \Longleftrightarrow \zeta([\Phi^1 - \Phi^2]) = 0
\end{equation}
That means for any $(i, j) \in E$
\begin{equation}
    (\Phi^1 - \Phi^2)_j - (\Phi^1 - \Phi^2)_i = 0
\end{equation}
Since we assume $G$ is connected, it indicates $\Phi^1 = \Phi^2 + c$, hence $[\Phi^1] = [\Phi^2]$.
As both $P^M$ and $T_\rho \Pcal(V)$ are linear space with dimension $M-1$, we prove $\zeta$ is a linear isomorphism.
\end{proof}

\subsection{Justification for Definition \ref{def:metric}}
Lemma \ref{lemma:isomorphism} gives an immersion \citep{do1992riemannian} from the tangent space $T_\rho \Pcal(V)$ to the the set of vector fields $F(G)$. Since we define the inner-product on $F(G)$ in \eqref{eq:flow_inner_product}, $\zeta$ naturally induce the metric on $T_\rho \Pcal(V)$. In this section, we will first justify $\langle \sigma^1, \sigma^2\rangle_\rho$ is valid.
Assume $\Phi^{\sigma^2}, \Psi^{\sigma^2} \in \zeta^{-1}(\sigma^2)$, then there exists $c$, such that $\Phi^{\sigma^2} = \Psi^{\sigma^2} + c$. Hence we have:
\begin{equation}
    \sum_{i=1}^M \sigma_i^1 (\Phi^{\sigma^2}_i - \Psi^{\sigma^2}_i) 
    = c \sum_{i=1}^M \sigma_i^1 = 0
\end{equation}
It shows that the value of $\langle \sigma^1, \sigma^2\rangle_\rho$ does not depends on the choice of the representative $\Phi^{\sigma^2}$, hence it is well-defined.

To show $\langle \sigma^1, \sigma^2\rangle_\rho$ is a valid inner-product, we need to check conjugate symmetry, linearity in the first argument, and positive-definiteness. For conjugate symmetry, we have:
\begin{align}
    \langle \sigma^2, \sigma^1\rangle_\rho
    &= \sum_{i=1}^M \sigma^2_i \Phi^{\sigma^1}_i \\
    &= \sum_{i=1}^M \div_\rho(\nabla \Phi^{\sigma^1})_i \Phi^{\sigma^1}_i \\
    &= \sum_{i=1}^M  \sum_{j \in N(i)} c_{ij}(\rho) (\Phi^{\sigma^1}_i - \Phi^{\sigma^1}_j) \Phi^{\sigma^2}_i \\
    &= \frac{1}{2} \sum_{i=1}^M  \sum_{j \in N(i)} c_{ij}(\rho) (\Phi^{\sigma^1}_i - \Phi^{\sigma^1}_j) \Phi^{\sigma^2}_i + \frac{1}{2} \sum_{j=1}^M  \sum_{i \in N(j)} c_{ij}(\rho) (\Phi^{\sigma^1}_i - \Phi^{\sigma^1}_j) \Phi^{\sigma^2}_i \\
    &= \frac{1}{2} \sum_{i=1}^M \Phi^{\sigma^2}_i \sum_{j \in N(i)} c_{ij}(\rho) (\Phi^{\sigma^1}_i - \Phi^{\sigma^1}_j) + \frac{1}{2} \sum_{j=1}^M \Phi^{\sigma^2}_j \sum_{i \in N(j)} c_{ij}(\rho) (\Phi^{\sigma^1}_i - \Phi^{\sigma^1}_j) \\
    &= \frac{1}{2} \sum_{i=1}^M \Phi^{\sigma^2}_i \sum_{j \in N(i)} c_{ij}(\rho) (\Phi^{\sigma^1}_i - \Phi^{\sigma^1}_j) + \frac{1}{2} \sum_{i=1}^M \Phi^{\sigma^2}_i \sum_{j \in N(i)} c_{ij}(\rho) (\Phi^{\sigma^1}_j - \Phi^{\sigma^1}_i) \\
    &= \frac{1}{2} \sum_{(i, j) \in E} c_{ij}(\rho)(\Phi^{\sigma^1}_i - \Phi^{\sigma^1}_j) (\Phi^{\sigma^2}_i - \Phi^{\sigma^2}_j) \label{eq:metric_phi}
\end{align}
We can see that \eqref{eq:metric_phi} does not depend on the order of $\sigma^1$ and $\sigma^2$, hence we have:
\begin{equation}
    \langle \sigma^1, \sigma^2 \rangle = \langle \sigma^2, \sigma^1 \rangle
\end{equation}
The linearity for the first argument is trivial to see. For positive-definiteness, if we have $\langle \sigma, \sigma \rangle_\rho = 0$, then by \eqref{eq:metric_phi}, we have:
\begin{equation}
    \sum_{(i, j) \in E} c_{ij}(\rho)(\Phi^{\sigma}_i - \Phi^{\sigma}_j)^2 = 0
\end{equation}
Since by our assumption, $G$ is connected and $c_{ij}(\rho) > 0$, it indicates $\sigma = \nabla \Phi^\sigma = 0$.
Finally, from \eqref{eq:metric_phi}, we can see that:
\begin{equation}
    \langle \sigma^1, \sigma^2 \rangle_\rho = \langle \nabla \Phi^{\sigma^1}, \nabla \Phi^{\sigma^2} \rangle_\rho
\end{equation}
This means the inner-product we defined in \eqref{eq:metric} is compatible with the immersion $\zeta$.

\subsection{Proof for Theorem~\ref{thm:main}}
We prove the theorem in a more general form in terms of free energy $F: \Pcal_2(V) \rightarrow \mathbb{R}$. Once we find the gradient flow for $F$, theorem \ref{thm:main} can be seen as a special case. In particular, when we define
\begin{equation}
    F(\rho) = \sum_{i=1}^M \rho_i f_i - \sum_{i=1}^M \rho_i \log \rho_i \label{eq:free_energy}
\end{equation}
we have $F(\rho) = D_\text{KL}(\rho || \pi)$.
\begin{proof}
The gradient flow in terms of the free energy is:
\begin{align}
    \frac{d \rho}{dt} = - \nabla_\rho F(\rho) \label{eq:gd}
\end{align}
By lemma \ref{lemma:isomorphism}, for any $\sigma \in T_\rho \Pcal_2(V)$, we have $\Phi^\sigma \in \zeta^{-1}(\sigma)$ such that $\sigma = \div_\rho(\Phi^\sigma)$. On the left hand side, we have:
\begin{equation}
    \langle \frac{d\rho}{dt}, \sigma \rangle = \sum_{i=1}^M \frac{d\rho_i}{dt} \Phi^\sigma_i 
\end{equation}
On the right hand side, we have:
\begin{align}
    \langle \nabla_\rho F(\rho), \sigma \rangle 
    &= \sum_{i=1}^M \frac{\partial F(\rho)}{\partial \rho_i} \sum_{j \in N(i)} c_{ij}(\rho) (\Phi^\sigma_i - \Phi^\sigma_j) \\
    &= \sum_{i=1}^M \frac{\partial F(\rho)}{\partial \rho_i} \sum_{j\in N(i)} c_{ij}(\rho) \Phi^\sigma_i - \sum_{i=1}^M \frac{\partial F(\rho)}{\partial \rho_i} \sum_{j\in N(i)} c_{ij}(\rho) \Phi^\sigma_j \\
    &= \sum_{i=1}^M \Phi^\sigma_i \sum_{j \in N(i)} c_{ij}(\rho) \frac{\partial F(\rho)}{\partial \rho_i} - \sum_{j=1}^M \Phi^\sigma_j \sum_{i\in N(j)} c_{ij}(\rho) \frac{\partial F(\rho)}{\partial \rho_i} \\
    &= \sum_{i=1}^M \Phi^\sigma_i \sum_{j \in N(i)} c_{ij}(\rho) \frac{\partial F(\rho)}{\partial \rho_i} - \sum_{i=1}^M \Phi^\sigma_i \sum_{j\in N(i)} c_{ij}(\rho) \frac{\partial F(\rho)}{\partial \rho_j} \\
    &= \sum_{i=1}^m \left(\sum_{j \in N(i)} c_{ij}(\rho) \big(\frac{\partial F(\rho)}{\partial \rho_i} - \frac{\partial F(\rho)}{\partial \rho_j}\big)\right) \Phi^\sigma_i
\end{align}
Hence we have:
\begin{equation}
    \sum_{i=1}^M \frac{d\rho_i}{dt} \Phi^\sigma_i = - \sum_{i=1}^m \left(\sum_{j \in N(i)} c_{ij}(\rho) \big(\frac{\partial F(\rho)}{\partial \rho_i} - \frac{\partial F(\rho)}{\partial \rho_j}\big)\right) \Phi^\sigma_i
\end{equation}
holds for arbitrary $\Phi^\sigma$. Then we prove:
\begin{equation}
    \frac{d\rho_i}{dt} = \sum_{j \in N(i)} c_{ij}(\rho) \left( \frac{\partial F(\rho)}{\partial \rho_j} - \frac{\partial F(\rho)}{\partial \rho_i} \right)
\end{equation}
Plug in the value of free energy in \eqref{eq:free_energy}, we have:
\begin{equation}
    \frac{d\rho_i}{dt} = \sum_{j \in N(i)} c_{ij}(\rho) [f_j + \log \rho_j - f_i - \log \rho_i]
\end{equation}
Thus we prove the theorem.
\end{proof}

\section{SAMPLER DETAILS}
We discuss different discretizations for DLD. Since we decompose the Markov jump process into independent sub-processes $X^t_n$, the distribution satisfies the following equation:
\begin{equation}
    \frac{d}{dt}\rho^t_n = \rho^t_n Q_n(x^t),
\end{equation}
where $x^t$ is the current state. The rate matrix $Q_n$ satisfies:
\begin{equation}
    Q_n(x^t)(i, j) = \left\{
    \begin{array}{cc}
    w_{ij}g\left(\frac{\pi(x^t_{\backslash n}, j)}{\pi(x^t_{\backslash n}, i)}\right),     & i \neq j  \\
    - \sum_{k \neq i}  w_{ik}g\left(\frac{\pi(x^t_{\backslash n}, k)}{\pi(x^t_{\backslash n}, i)}\right),   &  i= j
    \end{array}
    \right.
\end{equation}
Accordingly, the transition matrix satisfies:
\begin{equation}
    \frac{d}{dh} P_n^h(x^t) = P_n^h Q_n(x^t), \quad P_n^0(x^t) = I_C
\end{equation}
which has the following closed form solution:
\begin{equation}
    P_n^h(x^t) = \exp(\int_0^h Q_n(x^t) ds) = \exp(Q_n(x^t)h)
\end{equation}
To simplify the notation, we drop $x^t, n$, and only use $Q(i, j), P^h(i, j)$ if it does not cause ambiguity. 

\subsection{Single Jump}
\label{app:single}
We show that the first jump satisfies a categorical distribution. We define the holding time 
\begin{equation}
    S_i := \inf_t \{ t > 0 : X(0) = i, X(t) \neq i\}
\end{equation} as the time the process $X(t)$ stays at state $x$. We first show that $S_x$ is memoryless. That’s to say:
\begin{equation}
\mathbb{P}(S_i > r + t | S_i > t, X(0) = i) = \mathbb{P}(S_i > r+t | X(t) = i) =  \mathbb{P}(S_i > r | X(0) = i)    
\end{equation}
Since the only continuous memoryless distribution is exponential distribution, we know $S_x$ satisfies an exponential distribution $\lambda e^{-\lambda t}$. To estimate $\lambda$, we have:
\begin{align}
    \lambda & = - \frac{d}{dt}|_{t=0} e^{-\lambda t} = - \frac{d}{dt}|_{t=0} \mathbb{P}(S_i > t | X(0) = i) = \lim_{h \rightarrow 0} \frac{1 - \mathbb{P}(S_i > h | X(0) = i)}{h} \\
    &= \lim_{h \rightarrow 0} \frac{1 - \mathbb{P}(X(h) = i | X(0) = i) + o(h)}{h} = \lim_{h \rightarrow 0} \frac{1 - (1 + q_{ii}h + o(h))}{h} = - q_{ii}
\end{align}
To derive the transition probability, we condition on the holding time belongs to a small interval $(t, t+h]$, and let $h \rightarrow 0$, we have:
\begin{align}
    \tilde{P}_{ij} 
    &= \lim_{h\rightarrow 0} \mathbb{P}(X(t+h) = j | X(0) = i, t < S_i \le t+h) \\
    &= \lim_{h\rightarrow 0} \mathbb{P}(X(t+h) = j | X(t) = i, t < S_i \le t+h) \\
    &= \lim_{h\rightarrow 0} \frac{\mathbb{P}(X(t+h) = j | X(t) = i)}{\mathbb{P}(t < S_i \le t+h | X(t) = i)} \\
    &= \lim_{h \rightarrow 0} \frac{q_{ij} h + o(h)}{-q_{ii}h + o(h)} \\
    &= - \frac{q_{ij}}{q_{ii}}
\end{align}
Hence, the first jump that leaves state $i$ to $j \in N(i)$ satisfies the multinomial distribution $q(i, j) \propto w_{ij} g(\pi_j / \pi_i)$.

\subsection{DLMCf}
\label{app:dlmcf}
One of the most straightforward methods to estimate $P^h$ is forward Euler's method. Specifically, we can let:
\begin{equation}
    \tilde{P}_f^h = P^0 + h \frac{d}{dh} P^0 = P^0 + h Q
\end{equation}
where we use $f$ to indicate the method is based on forward Euler. The transition matrix can be written as:
\begin{equation}
    \tilde{P}_f^h = \left[
    \begin{array}{cccc}
    1 - h \sum_{j\neq 1} Q(1, j)  & h Q(1, 2) & \cdots & h Q(1, C)\\
    h Q(2, 1)   & 1 - h \sum_{j\neq 2} Q(2, j) & \cdots & h Q(2, n) \\
    \vdots & \vdots & \ddots  & \vdots \\
    h Q(n, 1)  & h Q(n, 2)  & \cdots & 1 - h \sum_{j\neq C} Q(C, j)\\
    \end{array} 
    \right] 
\end{equation}
One constraint we should take care of is that we need to restrict the simulation time $h$ such that the diagonal of $\tilde{P}^h$ is always non-negative.

\paragraph{Comparison with DMALA}
The DLMCf can be seen as a correction of DMALA \citep{zhang2022langevin}. Specifically, choosing $h = \exp(-\frac{1}{2})$, the transition matrix $\tilde{P}_\text{DMALA}^h$ of DMALA has the same off-diagonal value as $\tilde{P}_f^h$. However, the diagonal of the $\tilde{P}_\text{DMALA}^h$ is always $1$. This systematic mismatch reduces the efficiency DMALA. After correction, DLMCf has substantial improvements in efficiency. See more results in Appendix \ref{app:experiment}.

\subsection{DLMC}
\label{app:dlmc}
We denote the stationary distribution for $X^t_n$ as 
\begin{equation}
    \nu_n(x^t)(j) = \frac{\pi(x^t_{\backslash n}, j)}{\sum_{i \in \mathcal{C}} \pi(x^t_{\backslash n}, i)}
\end{equation}
Again, we drop $x^t$ and only use $\nu(j)$ if it does not cause ambiguity.
The transition matrix for \eqref{eq:P_interpolation} can be written as:
\begin{equation}
    \tilde{P}_n^h = \left[
    \begin{array}{cccc}
    \nu(1) + \sum_{j\neq 1} \nu(j) e^{-h \frac{Q(1, j)}{\nu(j)}}  & \nu(2) - \nu(2) e^{-h \frac{Q(1, 2)}{\nu(2)}} & \cdots & \nu(C) - \nu(C) e^{-h \frac{Q(1, C)}{\nu(C)}}\\
    \nu(1) - \nu(1) e^{-h \frac{Q(2, 1)}{\nu(1)}}   & \nu(2) + \sum_{j\neq 2} \nu(j) e^{-h \frac{Q(2, j)}{\nu(j)}}  & \cdots & \nu(C) - \nu(C) e^{-h \frac{Q(2, C)}{\nu(C)}}  \\
    \vdots & \vdots & \ddots  & \vdots \\
    \nu(1) - \nu(1) e^{-h \frac{Q(n, 1)}{\nu(1)}}   & \nu(2) - \nu(2) e^{-h \frac{Q(n, 2)}{\nu(2)}}  & \cdots & \nu(C) + \sum_{j\neq C} \nu(j) e^{-h \frac{Q(C, j)}{\nu(C)}}\\
    \end{array} 
    \right] 
\end{equation}
We can notice that when $C=2$, the estimation
\begin{equation}
    \tilde{P}^h = \left[
    \begin{array}{cc}
    \nu(1) + \nu(2) e^{-h \frac{Q(1, 2)}{\nu(2)}}  & \nu(2) - \nu(2) e^{-h \frac{Q(1, 2)}{\nu(2)}} \\
    \nu(1) - \nu(1) e^{-h \frac{Q(2, 1)}{\nu(1)}} & \nu(2) + \nu(1) e^{-h \frac{Q(2, 1)}{\nu(1)}}\\
    \end{array} 
    \right] = P_n^h
\end{equation}
is exact. For $C > 2$, We have:
\begin{equation}
    \tilde{P}_n^0 = I_C = P_n^0, \quad \tilde{P}_n^\infty = \nu^T 1 = P_n^\infty, \quad \frac{d}{dh}\tilde{P}_n^h|_{h=0} = Q_n = \frac{d}{dh}\tilde{P}_n^h|_{h=0}
\end{equation}

\paragraph{Comparison with PAS} The PAS \citep{sun2021path, sun2022optimal} flips a given number of sites $R$ per M-H step. As a results, it is equivalent with simulating the DLD via non-uniform time slice. Specifically, when the current state $x^t$ has small jump rate $Q$, to flip $R$ sites, PAS needs to simulate a longer time $h_+ > h$ in this M-H step. On the contrary, when the current state $x^t$ has large jump rate $Q$, PAS needs to simulate a shorter time $h_- < h$ in this M-H step. Consequently, the Markov chain obtained by PAS is more self correlated than DLMC. Also, since the PAS chain is likely to sample more frequently at the states with larger jump rates, M-H test need to reject more proposals to guarantee the chain is $\pi$-reversible. As a result, DLMC will be more efficient than PAS; see results in Appendix \ref{app:experiment}. One disadvantage of DLMC is that, the simulation time $h$ needed for transient and stationary phases are very different \citep{christensen2005scaling}, which makes tuning the scaling via average acceptance rate less robust comparing to PAS.

\subsection{Choice of Conductance}
\label{app:conductance}
Inspired by physics, we can define the conductance as
\begin{equation}
    c_{ij}(\rho) = \frac{m_{ij}(\rho) - m_{ji}(\rho)}{\log(m_{ij}(\rho)) - \log(m_{ji}(\rho))}, \ \forall j \neq i. \label{eq:conductance}
\end{equation}
The logarithmic mean in \eqref{eq:conductance} is known as \textit{conductance} in the stoichiometric network theory of chemical reactions \citep{qian2005thermodynamics}, where $m_{ij}$ represents the amount of the transition from $i$ to $j$, such that the numerator is the flux and the denominator is the driving force in nonequilibrium systems \citep{beard2007relationship}. We assume that the amount of the transition
\begin{equation}
    m_{ij}(\rho) = w_{ij} g(\frac{\pi_j}{\pi_i}) \rho_i, \ \forall j \neq i, \label{eq:transition}
\end{equation}
is only determined by the transition speed $w_{ij} g(\frac{\pi_j}{\pi_i})$ multiplying the current amount $\rho_i$. Here, $w_{ij}$ satisfies $w_{ij} = w_{ji}$ as an inherent scalar that measure the variability between $i$ and $j$, independent with both the target distribution $\pi$ and current distribution $\rho$. The second term $g(\frac{\pi_j}{\pi_i})$ is an external force caused by the target probability ratio and also relies on the choice of the weight function $g(\cdot)$. Furthermore, a reasonable assumption is that the transition should reach the equilibrium at the target distribution:
\begin{equation}
    m_{ij}(\pi) = m_{ji}(\pi) \label{eq:equilibrium}.
\end{equation}
Plug \eqref{eq:transition} into \eqref{eq:equilibrium}, one can solve
\begin{equation}
    g(\frac{\pi_j}{\pi_i}) = \frac{\pi_j}{\pi_i} g(\frac{\pi_i}{\pi_j}) \ \Rightarrow \ g(t) = t g(\frac{1}{t}), t > 0 \label{eq:g},
\end{equation}
which is exactly the \textit{locally balanced} (LB) function used in recent locally balanced samplers \citep{zanella2020informed}. Plug \eqref{eq:transition} and \eqref{eq:g} in \eqref{eq:conductance}, one can rewrite the conductance as:
\begin{equation}
    c_{ij}(\rho) = w_{ij}\frac{g(\pi_j / \pi_i)\rho_i - g(\pi_i/\pi_j) \rho_j}{f_i + \log\rho_i - f_j - \log \rho_j}.
\end{equation}

\section{EXPERIMENTAL DETAILS}
\label{app:experiment}
We focus on discrete spaces of the form $V = \mathcal{X}^D$ where $\mathcal{X} = \{e_1, ..., e_n\}$ is a finite set of one-hot vectors. We evaluate our methods on Bernoulli model, Ising model, factorial hidden Markov model and restricted Boltzmann machine. For each model, we consider both binary and categorical versions. For binary model, we use one high temperature setting and one low temperature setting. For categorical model, we use $n=4$ and $n=8$. We report the detailed descriptions of the models and corresponding results in the following. The running time for all methods across all models are summarized in Table \ref{tab:time}.
\begin{table}[h]
        \centering
        \caption{Running time (second) for all samplers on all target distributions with 100k steps}
        \begin{tabular}{c|cccccccc}
        \toprule
        Mehtod & hb-10-1 & bg-2 & rwm & gwg & dmala & pas & dlmc & dlmcf \\
        \midrule
Bernoulli (low) & 144& 61& 76& 231& 153& 406& 213& 149\\
Bernoulli (high) & 143& 61& 73& 197& 156& 245& 217& 150\\
Bernoulli ($n=4$) & 351& 112& 500& 115& 150& 514& 184& 170\\
Bernoulli ($n=8$) & 794& 458& 501& 121& 161& 526& 231& 194\\
Ising (low) & 203& 101& 149& 491& 490& 557& 548& 476\\
Ising (high) & 205& 106& 158& 519& 514& 589& 584& 507\\
Potts ($n=4$) & 484& 198& 412& 409& 452& 824& 494& 476\\
Potts ($n=8$) & 1335& 1022& 428& 416& 451& 832& 486& 470\\
binFHMM (low) & 216& 141& 228& 490& 476& 555& 542& 469\\
binFHMM (high) & 216& 141& 228& 495& 475& 559& 549& 474\\
catFHMM ($n=4$) & 450& 204& 492& 398& 433& 800& 470& 450\\
catFHMM ($n=8$) & 1490& 1168& 499& 393& 436& 804& 475& 456\\
binRBM & 144& 83& 105& 225& 235& 304& 296& 229\\
binRBM & 142& 82& 102& 229& 234& 305& 298& 229\\
catRBM ($n=4$) & 783& 357& 369& 195& 236& 590& 269& 255\\
catRBM ($n=8$) & 2721& 2283& 389& 276& 304& 684& 322& 313\\
        \bottomrule
        \end{tabular}
        \label{tab:time}
    \end{table}

\subsection{Bernoulli Model}
\begin{figure}[htb]
    \centering
    \includegraphics[width=.7\textwidth]{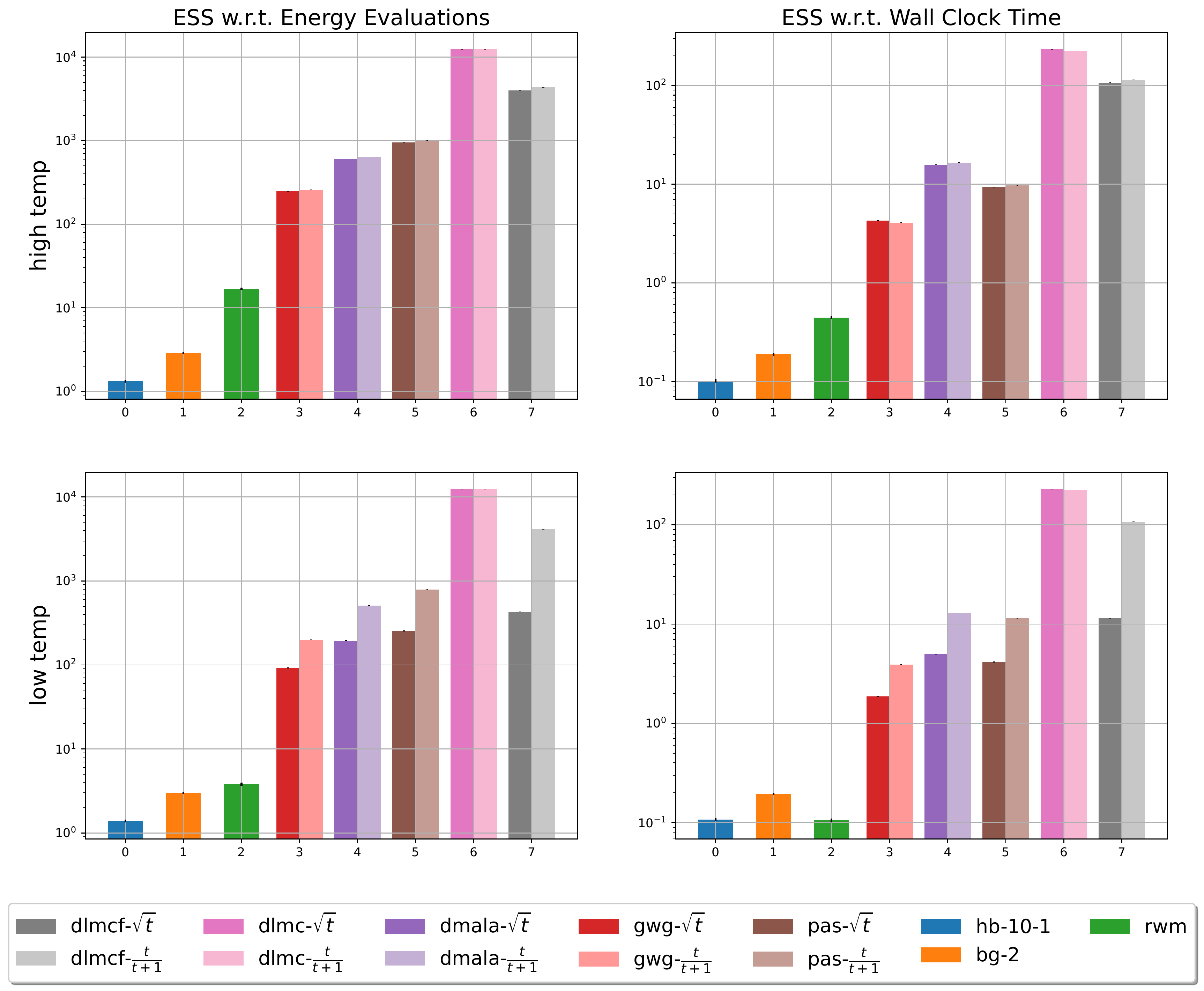}
    \caption{Evaluation on Bernoulli Models}
    \label{fig:bernoulli}
\end{figure}

\begin{figure}[htb]
    \centering
    \includegraphics[width=.7\textwidth]{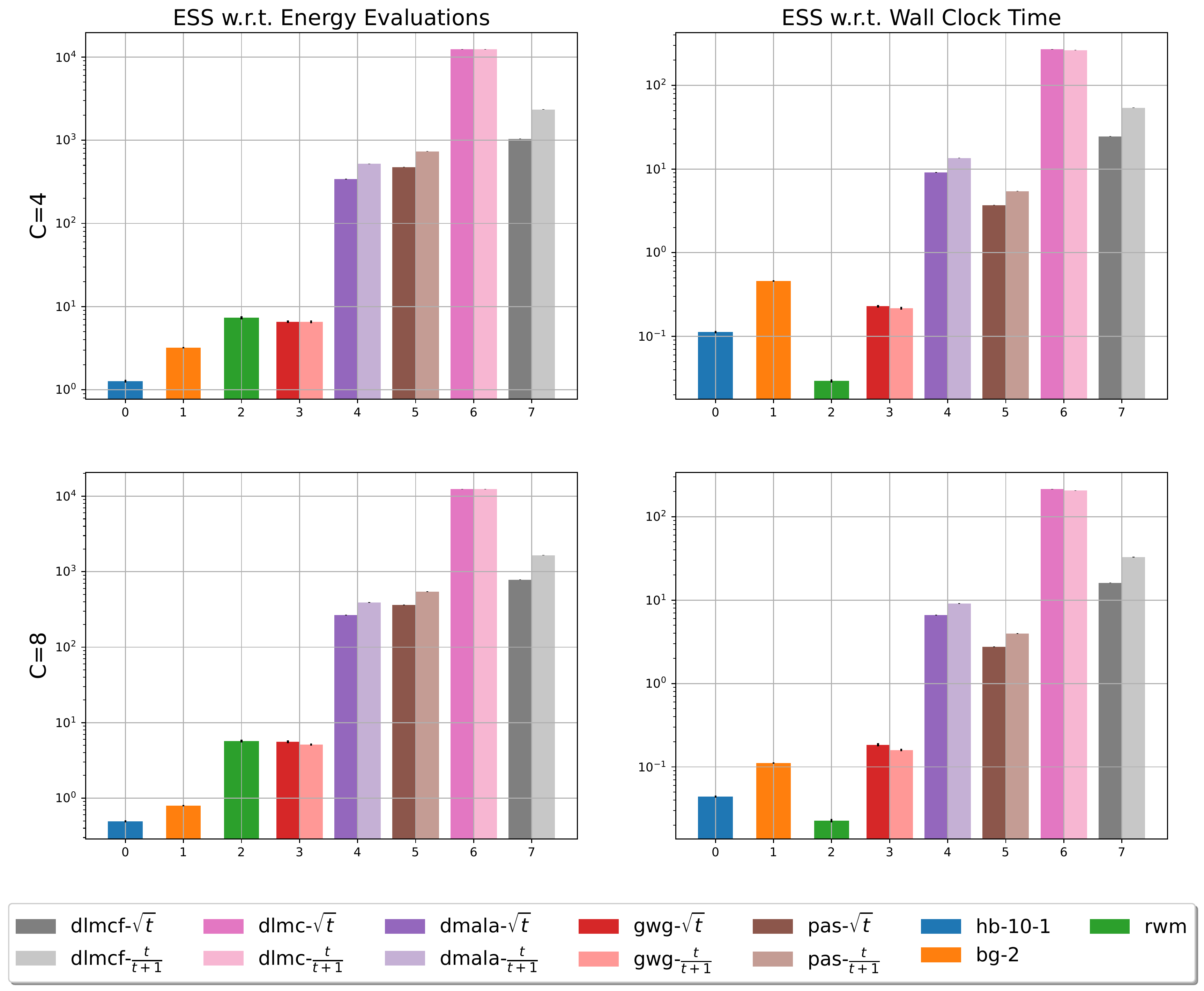}
    \caption{Evaluation on Categorical Models}
    \label{fig:categorical}
\end{figure}

The Bernoulli distribution is the simplest distribution in a discrete space, where each site is independent with others. For $x \in \mathcal{C}^N$, the energy function is:
\begin{equation}
    f(x) = \sum_{n=1}^N \langle x_n, \theta^d\rangle
\end{equation}
where $\theta^d \in \mathbb{R}^C$. Across all settings, we use the entries in $\theta_n$ independently sampled from centered normal distribution $\mathcal{N}(0, \sigma^2)$. For binary model we consider $D=10000$. We use $\sigma^2 = 0.125$ in the high temperature setting and $\sigma^2 = 12.5$ in the low temperature setting. For categorical model, we consider $D=2000$. We use $\sigma^2 = 1.125$ for both $n=4$ and $n=8$. The results are reported in Figure \ref{fig:bernoulli} and Figure \ref{fig:categorical}. We can see that DLMC and DLMCf have substantial better efficiencies compared to other samplers. The weight function $g(t) = \frac{t}{t+1}$ has better performance compared to $g(t) = \sqrt{t}$ as proved in \citet{zanella2020informed}. Moreover, the advantage of $g(t) = \frac{t}{t+1}$ is more significant when the target distributions are sharper, which is consistent with the observation in continuous space
\citep{livingstone2019barker}.

\subsection{Ising Model}
\begin{figure}[htb]
    \centering
    \includegraphics[width=.7\textwidth]{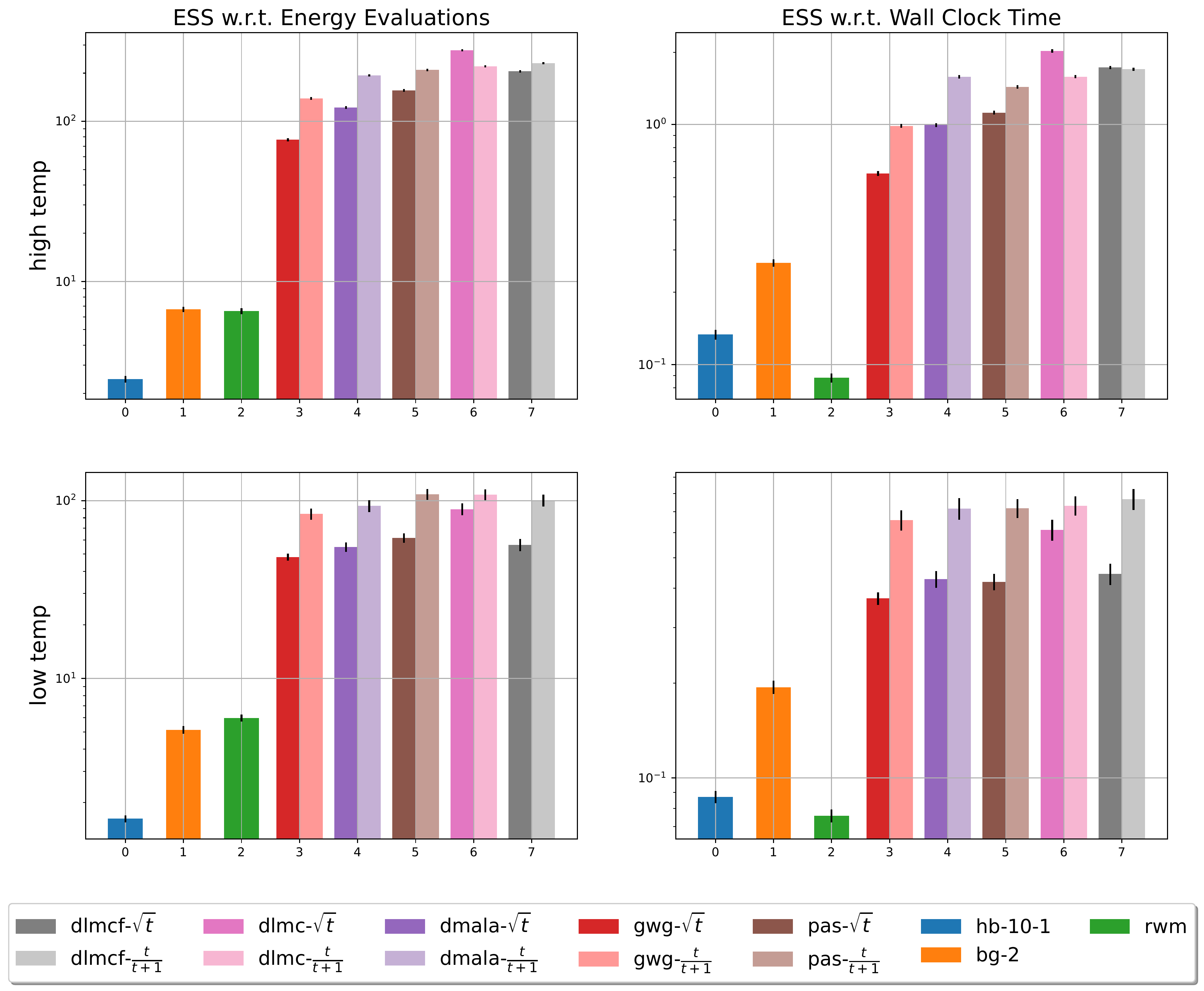}
    \caption{Evaluation on Ising Models}
    \label{fig:ising}
\end{figure}

\begin{figure}[htb]
    \centering
    \includegraphics[width=.7\textwidth]{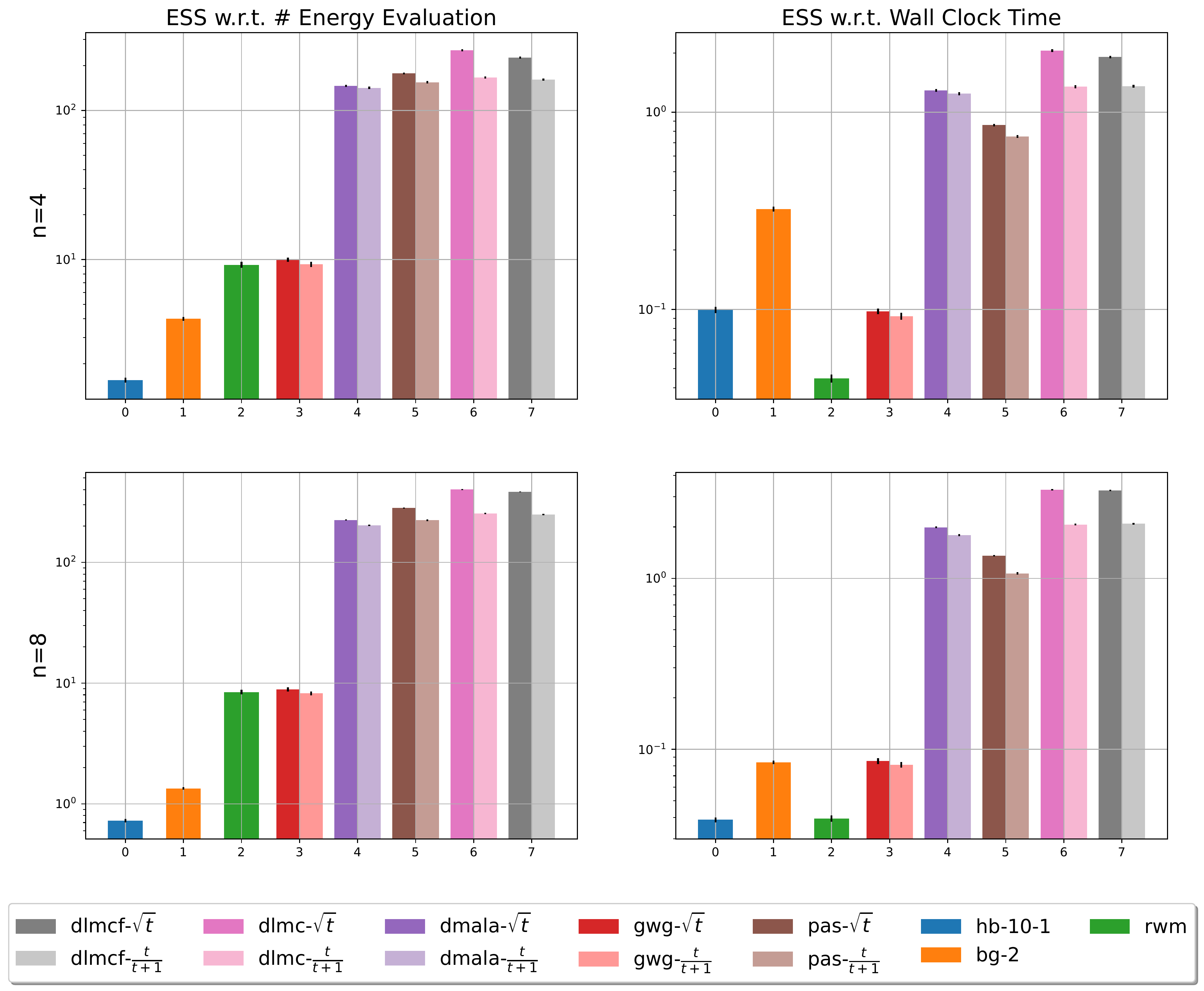}
    \caption{Evaluation on Potts Models}
    \label{fig:potts}
\end{figure}

The Ising model \citep{ising1924beitrag} is a mathematical model of ferromagnetism in statistical mechanics. It consists of binary random variables arranged in a lattice graph $G=(V, E)$ and allows node to interact with its neighbors. The Potts model \citep{potts1952some} is a generalization of the Ising model where the random variables are categorical. The energy function for Ising model and Potts model can be described as:
\begin{equation}
    f(x) = - \sum_{n=1}^N \langle x_n, \theta_n \rangle - \lambda \sum_{(i, j) \in E} \delta(x_i, x_j)
\end{equation}
where $\theta^d \in \mathbb{R}^n$, $\delta(x, y) = 1_{\{x = y\}}$. For Ising model, we consider $N=2500$ where $G$ is a $50 \times 50$ square lattice, and we follow the settings in \citet{zanella2020informed}. In high temperature setting, we use $\theta^d \sim \text{uniform}(-2, 1)$ for the outer part of the lattice graph, and $\theta^d \sim \text{uniform}(-1, 2)$ for the inner part of the lattice graph. The connection strength is chosen as $\lambda = 0.5$. In low temperature setting, we use $\theta^d \sim \text{uniform}(-4, 2)$ for the outer part of the lattice graph, and $\theta^d \sim \text{uniform}(-2, 4)$ for the inner part of the lattice graph. The connection strength is chosen as $\lambda = 1.0$. For potts model, we consider $N=900$ where $G$ is a $30 \times 30$ square lattice. For both $C=4, 8$, we use entries in external field $\theta^d_i \sim \text{uniform}(-1.5, 1.5) - 0.5 \frac{i}{C}$ for the outer part of the lattice graph, and $\theta^d_i \sim \text{uniform}(-1.5, 1.5) + 0.5 \frac{i}{C}$ for the inner part of the lattice graph, where $i=1, ..., C$. The connection strength is chosen as $\lambda = 1.0$. The results are reported in Figure \ref{fig:ising} and Figure \ref{fig:potts}. We can see that all LB samplers exhibit good performance. Among them, DLMC and DLMCf are the most efficient. The weight functions $g(t) = \sqrt{t}$ and $g(t) = \frac{t}{1+t}$ each demonstrate advantages for different samplers.

\subsection{Factorial Hidden Markov Model}
\begin{figure}[htb]
    \centering
    \includegraphics[width=.7\textwidth]{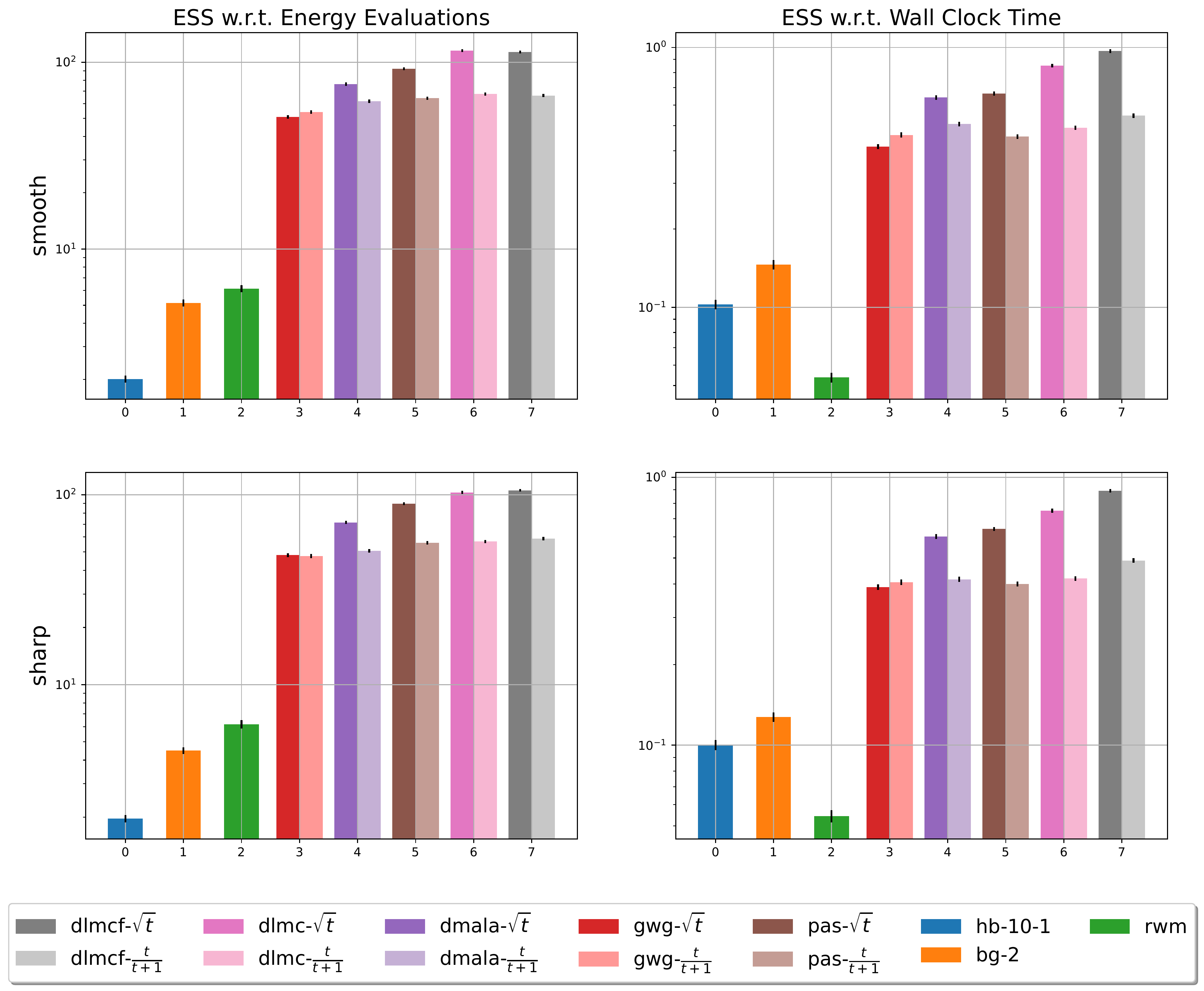}
    \caption{Evaluation on binFHMM}
    \label{fig:binFHMM}
\end{figure}

\begin{figure}[htb]
    \centering
    \includegraphics[width=.7\textwidth]{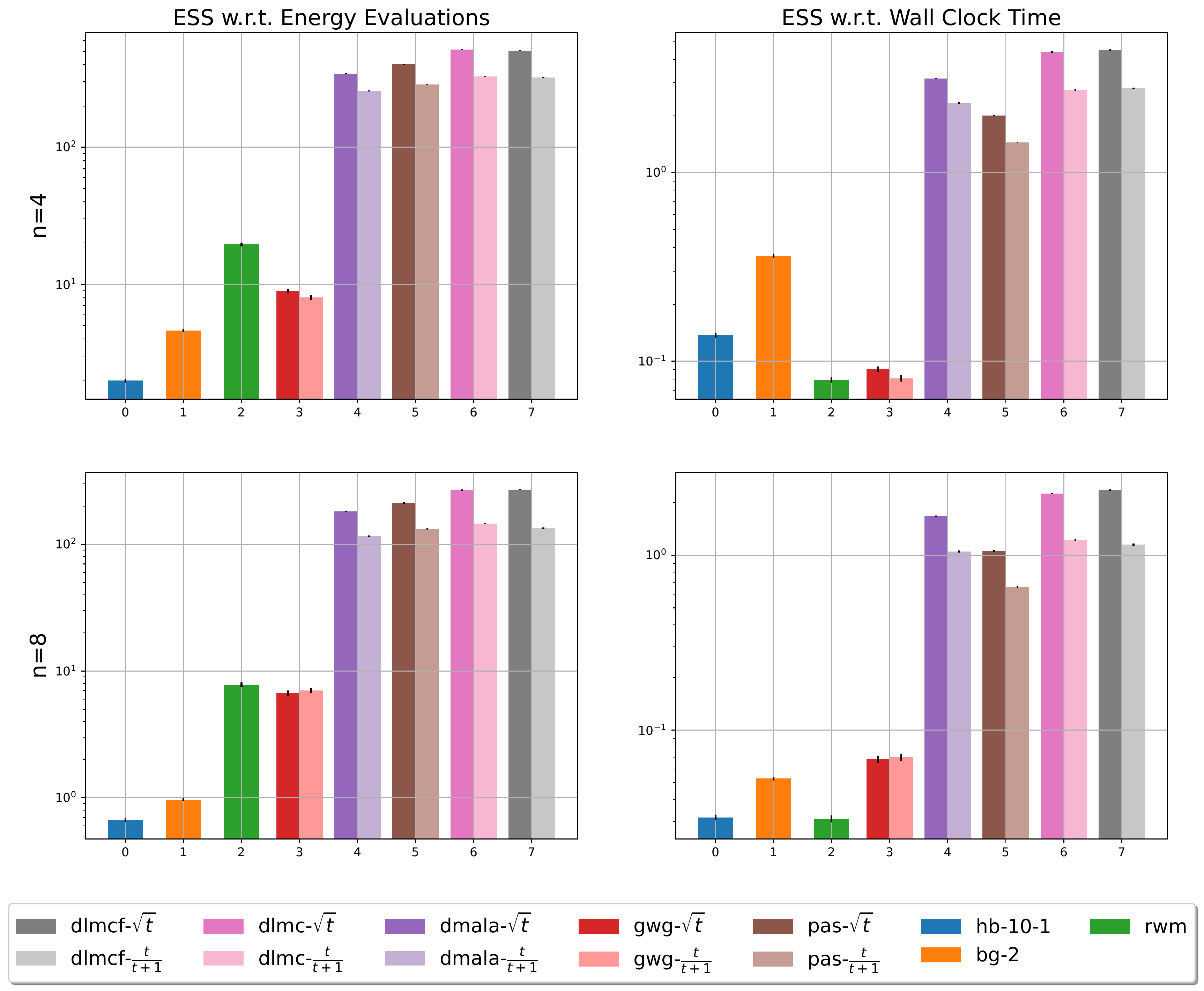}
    \caption{Evaluation on catFHMM}
    \label{fig:catFHMM}
\end{figure}

FHMM \citep{ghahramani1995factorial} uses latent variables to characterize time series data. In particular, it assumes the continuous data $y \in \mathbb{R}^L$ is generated by hidden state $x \in \mathcal{C}^{L\times K}$. The probability function is:
\begin{equation}
    p(x) = p(x_1) \prod_{l=2}^L p(x^t|x^{t-1}), \quad p(y|x) = \prod_{l=1}^L \mathcal{N}(y_t; \sum_{k=1}^K \langle W_k, x_{l, k}\rangle + b; \sigma^2)
\end{equation}
In particular, for binary model, we consider $\mathbb{P}(x_1=0) = 0.9, \mathbb{P}(x^t=x^{t-1}|x^{t-1}) = 0.8$, $\sigma = 2.0$. We use $L=200, K=10$ for high temperature setting and $L=100, K=20$ in low temperature setting. For categorical model, we use $p(x_1 | x_1\neq 0)$ and $p(x^t|x^{t-1}, x^t \neq x^{t-1})$ as uniform distribution and we use $L=200$, $K=10$.
We report the results in \figref{fig:binFHMM} and \figref{fig:catFHMM}.
Similar to the Ising model, we can see that all locally balanced samplers demonstrate good performance. In FHMM, LBP has an efficiency very close to the DLMC samplers. We believe this is because the energy change rate is stable in FHMM and the magnitude of the gradient changes steadily. Hence the Hamming distance works as a good metric. We also note that the weight function $g(t) = \sqrt{t}$ is systematically better than $g(t) = \frac{t}{t+1}$ on FHMM. This is consistent with the observation in \citet{livingstone2019barker} that $g(t) = \sqrt{t}$ performs better on smooth target distributions and $g(t) = \frac{t}{t+1}$ performs better on nonsmooth target distributions, although \citet{livingstone2019barker} focus on the sampling in continuous spaces.

\subsection{Restricted Boltzmann Machine}
\begin{figure}[htb]
    \centering
    \includegraphics[width=.7\textwidth]{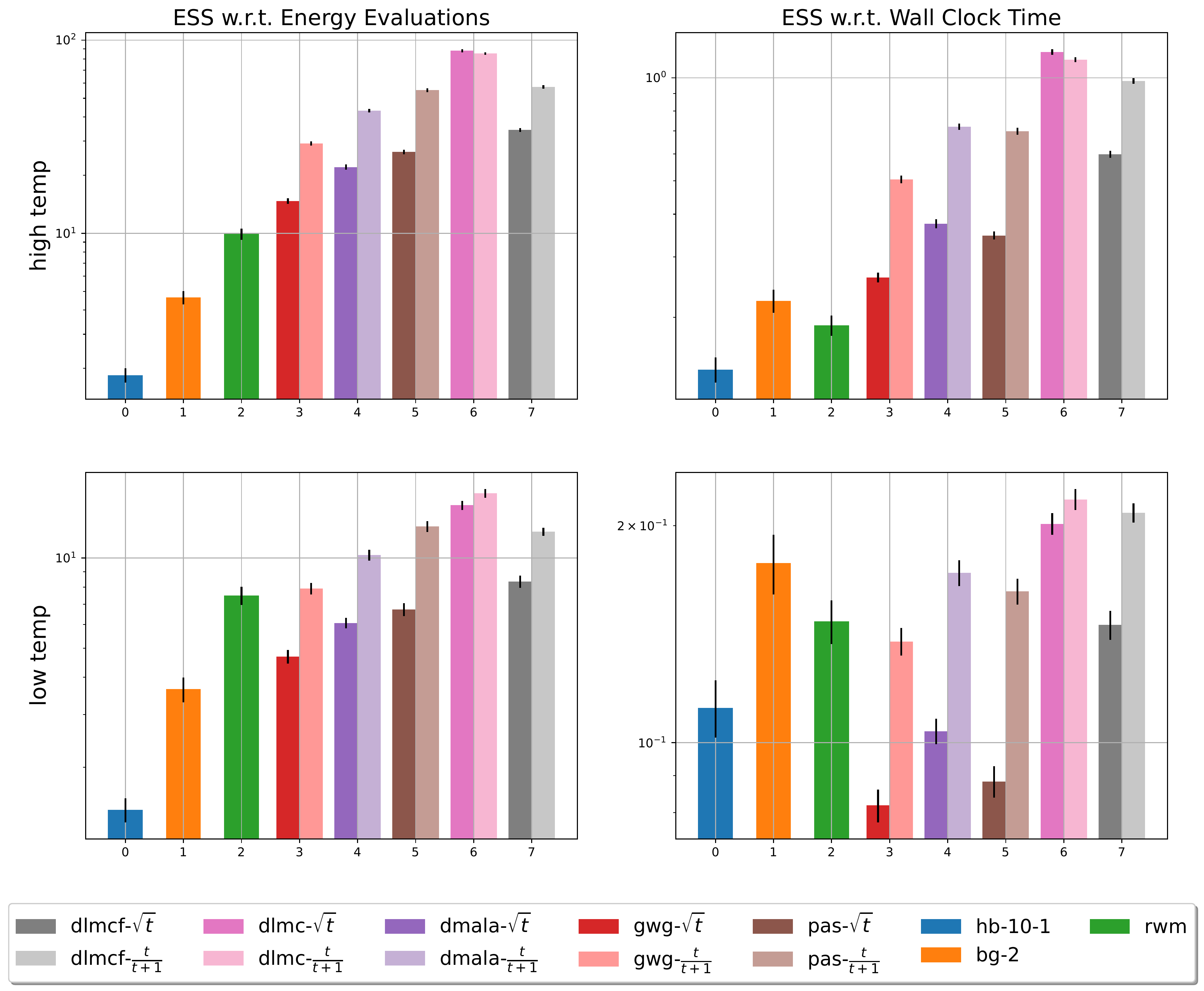}
    \caption{Evaluation on binRBM}
    \label{fig:binRBM}
\end{figure}

\begin{figure}[htb]
    \centering
    \includegraphics[width=.7\textwidth]{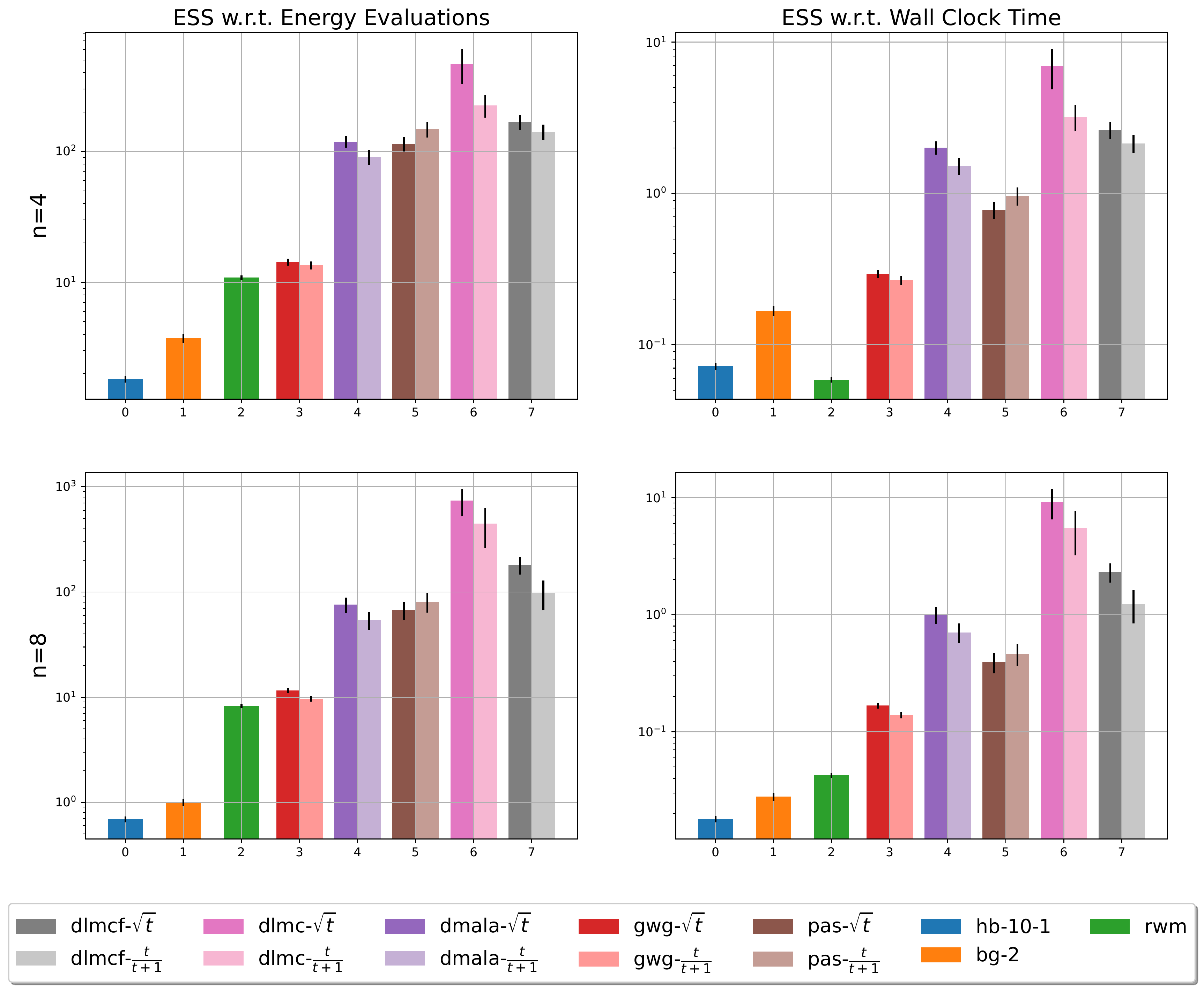}
    \caption{Evaluation on catRBM}
    \label{fig:catRBM}
\end{figure}

The RBM is an unnormalized latent variable model, with a visible random variable $v \in \mathcal{C}^N$ and a hidden random variable $h \in \{0, 1\}^M$. When $v$ is binary, we call it a binary RBM (binRBM) and when $v$ is categorical, we call it a categorical RBM (catRBM). The energy function of both binRBM and catRBM \citep{tran2011mixed} can be written as:
\begin{equation}
    f(v) = \sum_h \left[- \sum_{n=1}^N \langle v_n, \theta_n\rangle - \sum_{m=1}^M \beta_m h_m - \sum_{d, m} \langle h_m \theta_{m, d}, v_n \rangle\right]
\end{equation}
Unlike the previous three models, where the parameters are hand designed, we train binary RBM on MNIST \citep{lecun1998gradient} and categorical RBM on Fashion-MNIST \citep{xiao2017/online} using contrastive divergence \cite{hinton2002training}. Across all settings, we have $D=784$. For binary models, we use $M=25$ for high temperature setting and $M=200$ for low temperature setting. For categorical models, we use $M=100$. We report the results in \figref{fig:binRBM} and \figref{fig:catRBM}. The learned RBMs have stronger multi-modality compared to previous models. We can see that, as before, DLMC and DLMCf lead in proposal quality, while DLMC is the most efficient overall. 

\end{document}


%

%

\onecolumn
\aistatstitle{Instructions for Paper Submissions to AISTATS 2023: \\
Supplementary Materials}

\section{FORMATTING INSTRUCTIONS}